\documentclass[lettersize,journal]{IEEEtran}
\usepackage{cite}
\usepackage{amsmath,amssymb,amsfonts}
\usepackage{graphicx}
\usepackage{textcomp}
\usepackage{xcolor}
\def\BibTeX{{\rm B\kern-.05em{\sc i\kern-.025em b}\kern-.08em
    T\kern-.1667em\lower.7ex\hbox{E}\kern-.125emX}}

\usepackage{xcolor}
\usepackage{colortbl}
\usepackage{url}
\usepackage{xurl}
\usepackage{booktabs}
\usepackage[ruled]{algorithm2e}
\usepackage{algpseudocode}
\usepackage{enumerate}
\usepackage[english]{babel}
\usepackage{blindtext}
\usepackage{amsmath}

\usepackage{amssymb}
\usepackage{xspace}
\usepackage{multirow}
\usepackage{threeparttable}
\usepackage{float}
\usepackage{flafter}
\usepackage{comment}
\usepackage{enumitem}
\usepackage{subfig}
\usepackage{graphicx}
\usepackage{lipsum}

\def\fig{Fig.\xspace}

\def\tab{Tab.\xspace}

\def\ie{{\textit{i.e.}\xspace}} 
\def\eg{{\textit{e.g.}\xspace}}

\def\etc{{\textit{etc}\xspace}}

\newcommand{\head}[1]{{\noindent \textbf{#1:}}}

\graphicspath{{./figures/}}
\graphicspath{{./pdf/}}
\DeclareGraphicsExtensions{.pdf,.jpeg,.png}

\def\sysname{\textsc{NeurIT}\xspace}
\def\modulename{\textsc{TF-BRT}\xspace}

\ifodd 1
\newcommand{\com}[1]{\textbf{\color{red}(COMMENT: #1)}} 
\newcommand{\todo}[1]{\textbf{{\color{orange}(TODO: #1)}}}
\newcommand{\unused}[1]{{\color{gray}#1}}
\newcommand{\sheng}[1]{\textbf{\color{olive}(Sheng: #1)}} 
\else

\newcommand{\com}[1]{}
\newcommand{\todo}[1]{}
\newcommand{\unused}[1]{}
\newcommand{\sheng}[1]{}

\fi

\begin{document}

\title{
\sysname: Pushing the Limit of Neural Inertial Tracking for Indoor Robotic IoT
}

\author{\IEEEauthorblockN{Xinzhe Zheng$^\dagger$, Sijie Ji$^\ddagger$, Yipeng Pan$^\dagger$, Kaiwen Zhang$^\mathsection$, Chenshu Wu$^\boxtimes$} \\
\IEEEauthorblockA{\textit{Department of Computer Science, The University of Hong Kong} \\
$^\dagger$\{zxz.krypton, pan.yp\}@outlook.com
$^\ddagger$sijieji@caltech.edu \\
$^\mathsection$zhangkev@connect.hku.hk
$^\boxtimes$chenshu@cs.hku.hk}
}

\markboth{IEEE Transactions on Mobile Computing,~Vol.~14, No.~8, August~2021}%
{Shell \MakeLowercase{\textit{et al.}}: A Sample Article Using IEEEtran.cls for IEEE Journals}


\maketitle

\begin{abstract}
Inertial tracking is vital for robotic IoT and has gained popularity thanks to the ubiquity of low-cost inertial measurement units and deep learning-powered tracking algorithms.
Existing works, however, have not fully utilized IMU measurements, particularly magnetometers, nor have they maximized the potential of deep learning to achieve the desired accuracy.
To address these limitations, we introduce \sysname, which elevates tracking accuracy to a new level. 
\sysname employs a Time-Frequency Block-recurrent Transformer (\modulename) at its core, combining both RNN and Transformer to learn representative features in both time and frequency domains.
To fully utilize IMU information, we strategically employ body-frame differentiation of magnetometers, considerably reducing the tracking error.
We implement \sysname on a customized robotic platform and conduct evaluation in various indoor environments.
Experimental results demonstrate that \sysname achieves a mere 1-meter tracking error over a 300-meter distance.
Notably, it significantly outperforms state-of-the-art baselines by 48.21\% on unseen data.
Moreover, \sysname demonstrates robustness in large urban complexes and performs comparably to the visual-inertial approach (Tango Phone) in vision-favored conditions while surpassing it in feature-sparse settings.
We believe \sysname takes an important step forward toward practical neural inertial tracking for ubiquitous and scalable tracking of robotic things. 
\sysname is open-sourced here: \url{https://github.com/aiot-lab/NeurIT}.
\end{abstract}

\begin{IEEEkeywords}
neural inertial tracking, indoor localization, inertial measurement unit, time-frequency learning.
\end{IEEEkeywords}

\section{Introduction}
\label{sec:intro}
\IEEEPARstart{I}{ndoor} tracking is vital for robots, as it allows them to provide location-based services and navigate in areas where satellite signals are weak or unavailable.
Numerous techniques have been investigated for this purpose.
Vision-based methods~\cite{zhao2020masked} perform well in good conditions but struggle in simple environments and poor lighting.
\begin{figure}[t]
  \begin{center}
  \includegraphics[width=0.48\textwidth]{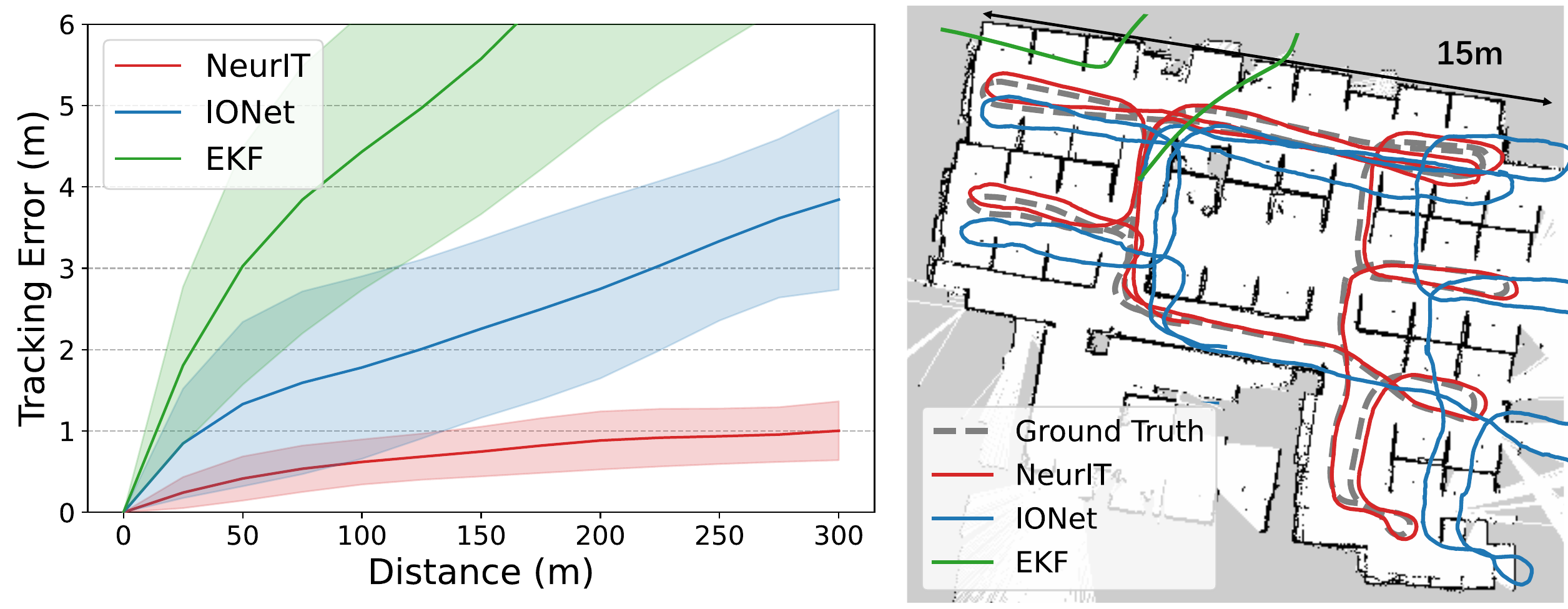}
  \caption{Tracking performance comparison and visualization of estimated trajectories.}
  \label{fig:intro_neurit}
  \end{center}
  \vspace{-5mm}
\end{figure}
LiDAR has been used to generate accurate tracking and mapping \cite{xu2021fast}, which, however, is too costly and may drift without feature points.
Wireless tracking solutions utilizing WiFi~\cite{wu2019rf}, millimeter-wave~\cite{rasekh2017noncoherent}, \etc, have also been widely studied, yet the performance is not robust in real-world cases, while the deployment overhead is significant.
These methodologies either rely on external infrastructure or depend on environmental features.
In contrast, inertial tracking using inertial measurement units (IMUs) has carved a niche by offering a truly infrastructure-free solution~\cite{ahmad2013reviews, zheng2025magnetometer}.
An IMU is a self-contained module integrating accelerometer, gyroscope, and magnetometer, which is a standard component in ubiquitous devices, including commercial robots.
The IMU-based tracking method does not depend on external infrastructure, rendering an inside-out solution that is suitable for indoor robotic navigation in various conditions.

However, achieving highly accurate inertial tracking is challenging due to sensor noise and drift.
Conventional studies examine filtering techniques like Kalman filter~\cite{ribeiro2004kalman}, complementary filter~\cite{euston2008complementary}, \etc, to reduce IMU noise.
These techniques may not work well in distinct conditions or for long durations.
The emergence of machine learning, specifically neural inertial tracking based on deep neural networks, offers a new approach that promises improved precision and generalizability.

Two types of neural networks are mainly exploited: recurrent neural networks (RNNs) and Transformers.
Despite acceptable results, their performance drops in cross-domain or unseen scenarios, indicating room for further improvement.
\begin{itemize}[leftmargin=*]
    \item RNN-based solutions can capture temporal features well~\cite{yan2019ronin, chen2018ionet, wei2019calibrating, gong2021robust}. However, they suffer from limited memory and operational inefficiencies, with issues such as vanishing gradients~\cite{gonzalez2018non} often destabilizing the learning process.
    \item Transformer-based approaches~\cite{rao2022ctin, wang2022a2dio} have shown promising results. Nevertheless, they struggle to capture information beyond the perception field, especially essential past data for the current motion context.
    \item Existing works mostly employ accelerations and gyroscopes only, and exclude the magnetometer for learning.
    Few works have integrated magnetometers for drift compensation, yet merely employ complementary filter-based techniques~\cite{shen2018closing,gong2021robust}, yielding only marginal enhancements in the outcomes.
\end{itemize}

To overcome these limitations, we present \sysname, a robust system that unleashes the full potential of neural inertial tracking and achieves remarkable accuracy for robotic tracking. 
\sysname excels mainly in two aspects:
First, based on an in-depth understanding of inertial data properties, \sysname incorporates magnetometer data as an additional feature with accelerometer and gyroscope measurements.
Although the magnetometer has previously been believed to be susceptible to indoor interference~\cite{chung2011indoor,fan2017accurate,ho2020using} and was excluded from learning, we noticed that it has a lower drift than the gyroscope in providing long-term orientation information, making it a valuable input for drift compensation.
It is noticeable that magnetometers primarily capture pose information, while accelerations and gyroscopes depict the movement at a fine-grained level.
To align these features together, we integrate \textit{magnetometer differentiation within the body frame}.
This ensemble enhances the network's ability to counteract the noise-induced drift in pose estimation.
Second, to maximize the potential of neural networks for enhanced inertial tracking, we propose \textit{Time-frequency Block-Recurrent Transformer}, named \modulename, which embraces the advantages of both RNN and Transformer and overcomes their respective drawbacks by extracting pre-sequence information and local motion information within the perception field.
Particularly, \modulename extracts both time and frequency domain information to obtain a more comprehensive understanding of object movement.
By learning from complete IMU data with an advanced neural network design, \sysname achieves robust tracking capabilities that endure extended duration and cover substantial distances.

To implement and evaluate \sysname, we build a customized robotic platform, which consists of a commodity 9-axis IMU for \sysname, a LiDAR-inertial system for ground truth, and a visual-inertial system for comparison.
We verify the performance of \sysname on two public benchmarks and a newly built dataset. 
Our dataset is collected in three different campus buildings, totaling 33.7 km over 15 hours.
The experimental results demonstrate that \sysname not only outperforms all baseline models in the existing benchmark dataset, but it also achieves better results under more challenging conditions.
Specifically, compared to the best baseline, \sysname improves the tracking accuracy by 48.21\% on unseen data and achieves an average drift rate of 0.62\%, indicating its superior robustness over long distances.
Additionally, \sysname achieves the thus far best performance on two benchmark datasets, RIDI~\cite{yan2018ridi} and RONIN~\cite{yan2019ronin}.
To evaluate its real-world applicability, we test \sysname in two large urban complexes, a shopping mall and a concert hall, where it maintains strong robustness in complex environments.
Further, when compared to a commercial visual-inertial system~\cite{nguyen2017assessing}, 
\sysname exceeds in plain environments while achieving comparable performance in visual-favored environments. 
\fig\ref{fig:intro_neurit} illustrates the superior tracking accuracy achieved by \sysname, while prior neural inertial tracking approach (IONet~\cite{chen2018ionet}) and traditional control method (EKF~\cite{ribeiro2004kalman}) fail to provide robust tracking outcomes.
This enhances the potential of inertial tracking as a viable solution for various indoor tracking applications, inviting new perspectives on the longstanding challenge of indoor positioning.

In summary, our core contributions are as follows: 
\begin{enumerate}[leftmargin=*]
    \item With in-depth insight into IMU, we integrate magnetometers for neural inertial tracking.
    We propose an effective sensor fusion approach by using the derivative of body-frame magnetometers, which demonstrates substantial enhancements in tracking accuracy and robustness.
    \item We propose \sysname, an inertial tracking system that fully utilizes IMU data under an advanced network named \modulename.
    \modulename combines the strengths of RNNs and Transformers to effectively learn time-frequency features, enabling accurate tracking across diverse environments while minimizing long-term drift.
    \item We implement \sysname on a customized robotic platform and test it in various environments using a newly built dataset. \sysname surpasses the existing baselines on all the benchmarks and performs well against a visual-inertial tracking system. We have open-sourced the collected dataset and source code for the research community.
\end{enumerate}

In the rest of the paper, we review the literature in \S\ref{sec:related_works}.
Then we investigate the insights of neural inertial tracking in \S\ref{sec:intuition} and present the system design in \S\ref{sec:design}.
Implementation details are given in \S\ref{sec:implementation}, followed by experiment settings in \S\ref{sec:exp}.
We evaluate \sysname scrupulously in \S\ref{sec:evaluation}.
Finally, we conclude the paper in \S\ref{sec:conclusion}.

\begin{figure}[t]
\centering
   \subfloat[Place A]{
    \label{fig:bb-magn}
      \includegraphics[width=0.45\columnwidth]{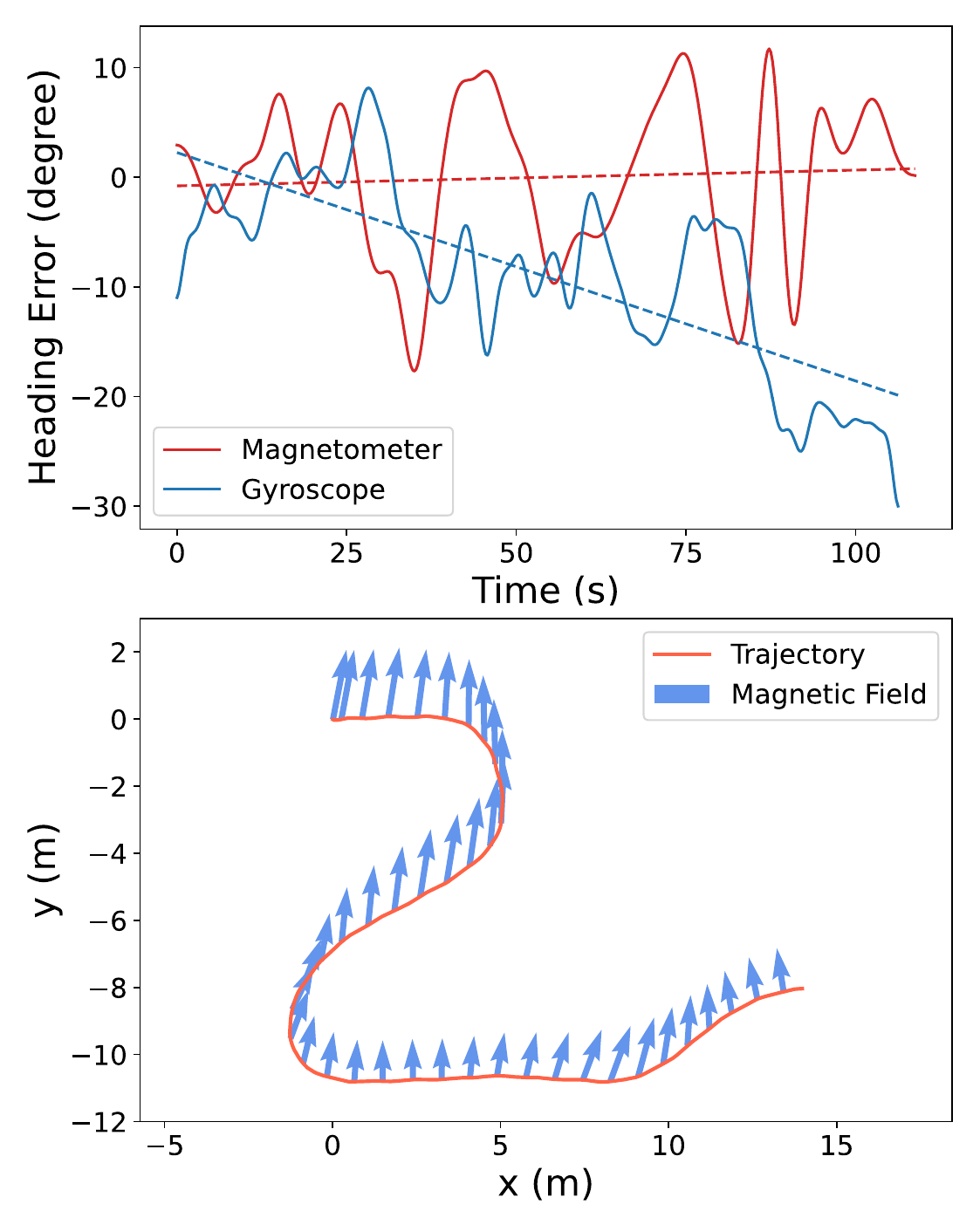}
      }
   \subfloat[Place B]{
    \label{fig:bc-magn}
      \includegraphics[width=0.45\columnwidth]{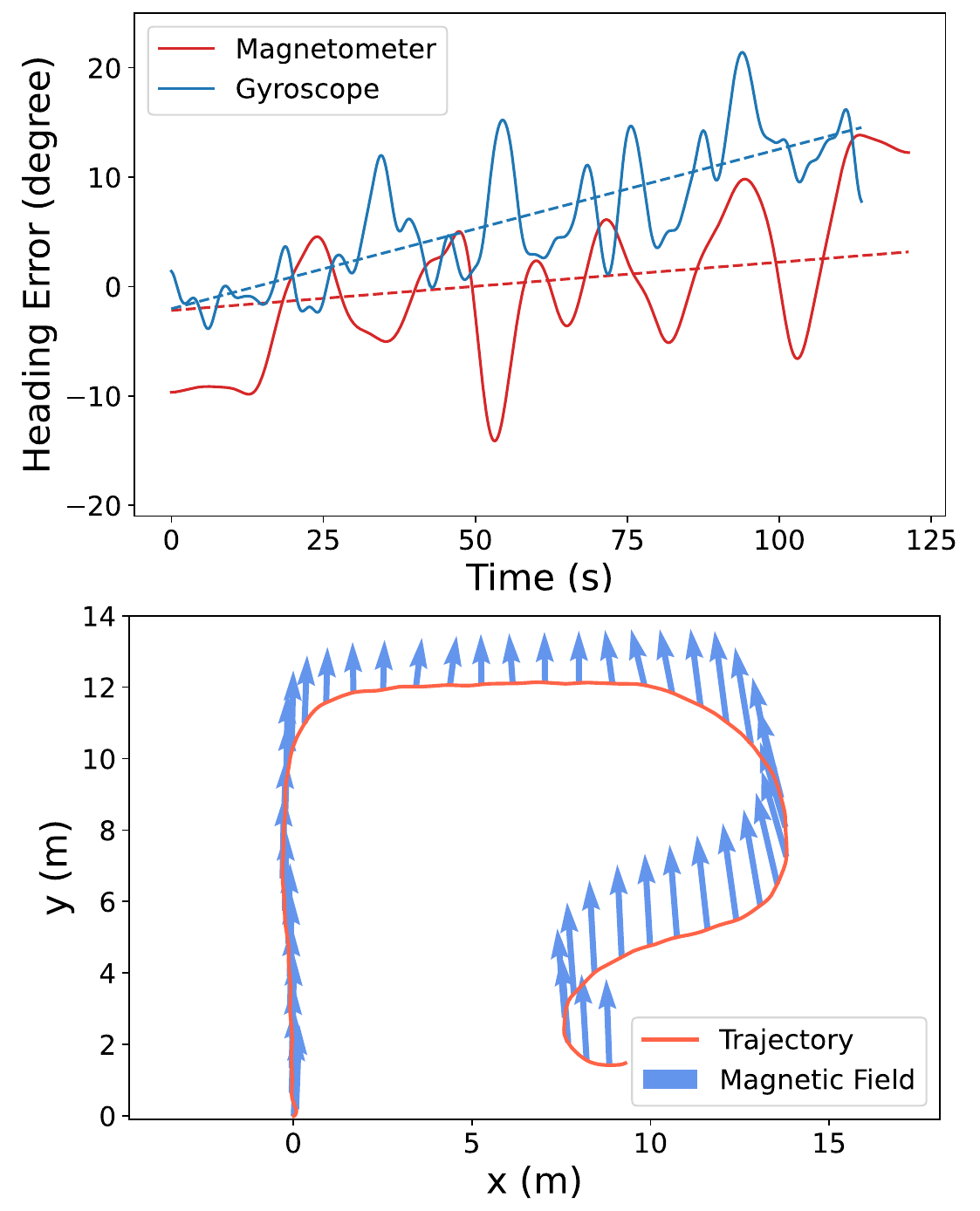}
      }
      \caption{Global magnetometers study.
      First row: heading error comparison between global magnetometers and gyroscopes; Second row: visualization of global magnetometers.}
   \label{fig:magn-vis}
\end{figure}

\section{Related Works}
\label{sec:related_works}
\head{Neural Inertial Tracking}
IONet~\cite{chen2018ionet} first integrates neural networks for inertial tracking, which greatly improves the accuracy compared with traditional methods~\cite{shen2018closing,gong2021robust}.
With its success, RONIN~\cite{yan2019ronin} and DeepVIP~\cite{zhou2022deepvip}, \etc, continue to explore the limit of RNN for inertial tracking and achieve promising results.
After that, CTIN~\cite{rao2022ctin}, A2DIO~\cite{wang2022a2dio}, \etc, replicate the success of the Transformer in the NLP field and further improve the tracking accuracy.
Additionally, researchers have developed models optimized for mobile deployment~\cite{liu2023smartphone}, while LLIO~\cite{wang2022llio} has been specifically designed for wheeled vehicles.
To the best of our knowledge, \sysname is the first work to combine the advantages of RNN and Transformer for neural inertial tracking.

\head{Magnetometers Usage}
Most aforementioned neural inertial tracking works use accelerations and gyroscopes exclusively since magnetometers will be affected by indoor environments~\cite{fan2017accurate}.
Some works, like $A^3$~\cite{zhou2014use} and MUSE~\cite{shen2018closing}, utilize magnetometers to calibrate gyroscope drift.
In addition, Tlio~\cite{liu2020tlio} incorporates deep learning to enhance the robustness of EKF.
However, the fundamental idea of these works is under the framework of filtering techniques, which is distinct from \sysname. 
\sysname utilizes \textit{differentiation of body-frame magnetometers} to achieve fusion and complementarity of IMU data from a neural network input perspective.

\head{Fused Inertial Tracking}
Despite the impressive capabilities of modern neural networks, accumulating path errors over time remains a challenge for precise indoor positioning. To address this, integrating indoor maps is essential for achieving high-precision localization.
For instance, NILoc~\cite{herath2022neural} employs a transformer-based architecture to determine device location from velocity sequences, while~\cite{melamed2022learnable} uses spatial map embeddings with IMU data for indoor location prediction.
Other approaches, such as combining IMU estimations with WiFi~\cite{wu2019easitrack,nurpeiissov2022end} or acoustic signals~\cite{wang2019acoustic}, offer limited improvements and are affected by ambient interference.
Visual~\cite{loo2019cnn,boikos2016semi} and LiDAR-based~\cite{zhang2014loam} tracking solutions have shown promise in both indoor and outdoor environments. 
However, these methods are prone to drift and tracking failure in feature-sparse areas. To enhance system robustness, researchers have incorporated IMU information, resulting in visual-inertial~\cite{eckenhoff2021mimc} and LiDAR-inertial~\cite{xu2021fast} tracking systems.

\section{Observations and Measurements}
\label{sec:intuition}

IMU-based tracking often suffers from orientation drift, primarily caused by gyroscope noise.
While global magnetometers can help correct this drift outdoors, indoor environments pose challenges due to magnetic distortions from ferromagnetic building materials~\cite{chung2011indoor,fan2017accurate}.
This distortion may explain why prior neural inertial tracking studies often exclude magnetometer inputs.
To test this hypothesis, we collected magnetometer and gyroscope data from different indoor environments to estimate the heading orientations.
We compared their accuracy by calculating the heading error over time, as shown in \fig\ref{fig:magn-vis}.
The first row presents heading errors from global magnetometers and gyroscopes.
Despite fluctuations and distortions caused by complex indoor structures, the dashed lines reveal that magnetometers exhibit significantly less long-term drift than gyroscopes.
This intriguing finding suggests that magnetometers may serve as a valuable complementary feature for neural inertial tracking.

Nevertheless, effectively integrating the three modalities remains challenging due to their distinct perspectives on motion sensing.
First, for precise floor plane tracking, converting accelerations and gyroscopes from the IMU body frame to the Earth frame is crucial since the IMU operates in a non-inertial coordinate system.
Unlike accelerometers and gyroscopes, which measure the amount of motion but not the precise direction, magnetometers can directly infer the orientation of an object.
To improve the alignment of magnetometers with accelerometers and gyroscopes, we introduce the \textit{derivative of magnetometers in the body frame}, which encompasses both the pose information and the object’s orientation motion, enhancing the network’s ability to learn motion patterns at a finer-grained level.
In \S\ref{sec:magn-study}, we present a study demonstrating that integrating magnetometer data significantly enhances the performance of our system.

\begin{figure}[t]
  \begin{center}
  \includegraphics[width=0.45\textwidth]{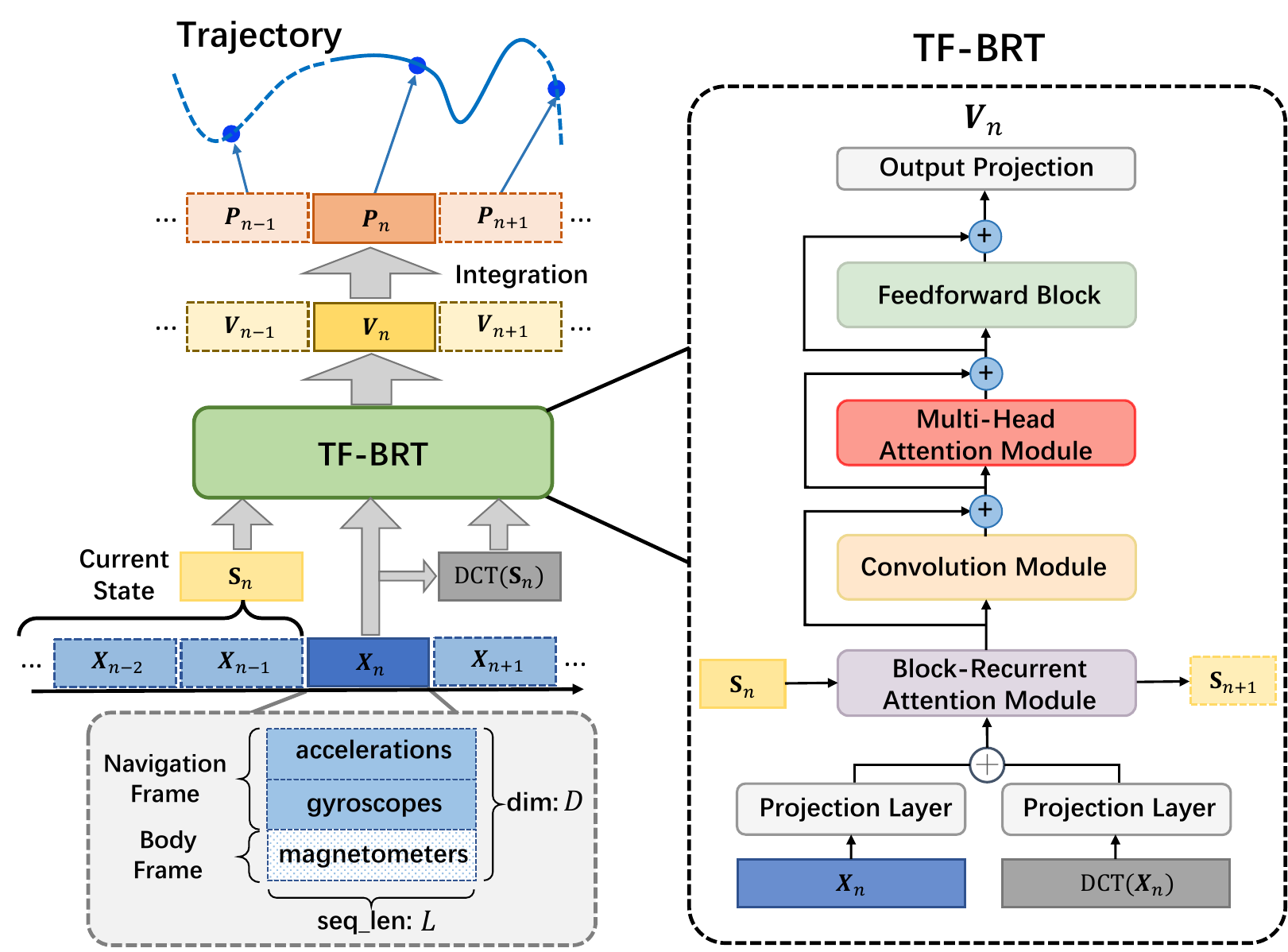}
  \caption{\sysname Overview. Left side: End-to-end indoor tracking framework; Right side: \modulename architecture.}\label{fig:framework-overview}
  \end{center}
  \vspace{-5mm}
\end{figure}

\section{\sysname Design}
\label{sec:design}

\subsection{Overview}
\label{overview}

\sysname presents a sequence-to-sequence framework to reconstruct the moving trajectories of robots using IMU sensors only, without relying on any external infrastructure (\eg, access points, cellular stations). 
\fig\ref{fig:framework-overview} overviews the design of \sysname.
To track the robot's movement, IMU readings that consist of accelerations, gyroscopes, and magnetometers are collected and pre-processed into a time series representation.
The core of \sysname is a neural network called \modulename, which leverages block-recurrent attention~\cite{hutchins2022block} and time-frequency learning.
The block-recurrent attention mechanism enables the model to leverage both current IMU data and historical context for more accurate tracking.
By incorporating features from both time and frequency domains, the model improves its ability to handle diverse motion states.
At the end of \sysname, \modulename outputs a sequence of velocities and orientations, which are integrated to compute the robot's location over time.
We avoid direct location prediction, as location is a continuous variable without a finite set.
Additionally, we introduce multi-loss learning and data augmentation to further enhance the robustness of \modulename.

\subsection{Data Preparation}
\label{data-preparation}
The IMU coordinate frame, which constantly changes, reports the raw IMU data.
The Earth coordinate frame, on the other hand, defines the moving trajectory. 
To resolve this discrepancy, a coordinate transformation is required.
A common practice in neural inertial tracking is to transform the raw IMU data and the ground truth into the same coordinate frame called the navigation frame~\cite{yan2019ronin}.
We use the same coordinate frame transformation as RONIN \cite{yan2019ronin} in \sysname, which uses a \textit{heading-agnostic coordinate frame} (Earth coordinate frame) to represent the IMU data.

The coordinate transformation is only applied to accelerations and gyroscopes: $(\textbf{a}^{g},\textbf{w}^{g})=\textbf{q}(\textbf{a}^{b}, \textbf{w}^{b})\textbf{q}^{*}$,
where $\textbf{a}$ and $\textbf{w}$ denote the accelerations and gyroscopes, $b$ and $g$ represent the body frame and Earth coordinate frame.
\textbf{q} is the quaternion, and $\textbf{q}^{*}$ is its conjugate, with both provided by the IMU.
Conversely, we keep the magnetometers in the body frame and apply differentiation operation ($\dot{\Box}$):
\begin{equation}
\label{eqn:magn-der}
\begin{aligned}
    \dot{\mathbf{m}}^{b}&=\dot{\mathbf{R}} \cdot \mathbf{m}^{g} \\
    &=\mathbf{\Omega} \cdot \mathbf{R} \cdot \mathbf{m}^{g} \\
    &=\begin{pmatrix}
    0&  -w_z& w_y\\
    w_z&  0& -w_x\\
    -w_y&  w_x& 0
    \end{pmatrix} \cdot \mathbf{R} \cdot \mathbf{m}^{g},
\end{aligned}
\end{equation}
where $\dot{\bold{m}}^{b}$ is the time derivative of the magnetometer vector in the body frame, $\bold{R}$ is the rotation matrix from the Earth frame to the body frame, and $\bold{\Omega}$ is the angular velocity matrix formed from the gyroscope readings.
Obviously, with the hidden rotation information, $\dot{\mathbf{m}}^{b}$ can be used to compensate for gyroscope drift as described in the aforementioned context.

The combined three-modality time-series data, denoted as $\textbf{X} = [\textbf{a}^g;\textbf{w}^g;\dot{\textbf{m}}^b]$, is segmented into fixed-length sequences of length $L$ using a sliding window.
These segments constitute the data samples for the model input $\textbf{X}_n \in \mathbb{R}^{L\times D}, n=[1,2,3,\cdots]$, where $D$ is the IMU data dimension.

\subsection{Model Architecture}
\label{model-architecture}
We construct \modulename based on advanced neural network architectures to improve the learning outcome for IMU-based robot tracking.
\modulename comprises six key components, as shown in \fig \ref{fig:framework-overview}: time-frequency input projection, block-recurrent attention~\cite{hutchins2022block} module, convolution~\cite{lecun1995convolutional} module, multi-head attention~\cite{vaswani2017attention} module, feedforward module, and output projection layer.
The time-frequency input projection captures relevant features from both domains, revealing latent patterns in the robot’s motion.
The block-recurrent attention module integrates RNNs into the Transformer, enabling pre-sequence modeling and contextual understanding within the receptive field for more accurate trajectory prediction.
To make the network convergence faster, the remaining four modules are interconnected in a residual way~\cite{gulati2020conformer} with a pre-norm residual unit~\cite{Dai2019TransformerXLAL}.
For the convolution module, a depthwise separable convolution~\cite{chollet2017xception} layer is sandwiched between the layernorm and dropout layer.
Depthwise separable convolution can efficiently capture a sequence's local features.
The multi-head attention module employs a multi-head self-attention layer positioned between the layernorm and dropout layers.
Finally, the feedforward module utilizes two fully-connected layers to extract the final features.
After all the processes are completed, the output projection layer effectively distills the robot's velocity information from the high-dimensional space, resulting in the velocity sequence $\textbf{V}_n$.
This design approach excels in handling long-time sequences and shows promising generalization abilities in various environments.

\subsubsection{Time-Frequency Input Projection}
\label{tf-input-projection}
Recent studies on time series prediction and classification have emphasized the importance of incorporating frequency domain information.
It has been suggested that the distinct patterns of various time series can be expressed through the frequency domain for an enhanced performance~\cite{li2021units, ding2020rf, yao2019stfnets}.
One such approach, UniTS~\cite{li2021units}, improves IMU motion recognition by using Short-Time Fourier Transform (STFT) to initialize the convolution kernels, surpassing models that rely only on time-domain data.

Inspired by prior research, \modulename utilizes dual-channel inputs consisting of both time and frequency domain features.
However, unlike UniTS and RF-Net~\cite{ding2020rf}, which use STFT and FFT as separate methods to extract frequency features, \modulename utilizes the Discrete Cosine Transform (DCT) to extract frequency domain information, avoiding the need for complex number operations in the network.

The two input channels are connected by projection layers, which extract hidden features in the two channels through the fully connected layer and map the IMU feature domain ($\mathbb{R}^{L\times D}$) to a larger domain ($\mathbb{R}^{L\times D_h}$), enabling subsequent networks to effectively learn and operate within the high-dimensional feature space.
$D_h$ represents the feature dimension in the following modules.
The data after the projection layers are merged to combine the extracted high-dimensional features of the time domain and frequency domain.
This operation can be written as $f_{\theta_t}(\textbf{X}_n)+f_{\theta_f}({\rm{DCT}}(\textbf{X}_n))=\textbf{M}_n$, where $\theta_t$ and $\theta_f$ denote the projection layer parameters of time channel and frequency channel, respectively, and $\textbf{M}_n$ represents the merged hidden features.

\begin{figure}[t]
\centering
   \subfloat[Workflow of $\textbf{E}_{n+1}$]{%
    \label{fig:workflow_en1}
      \includegraphics[width=0.42\columnwidth]{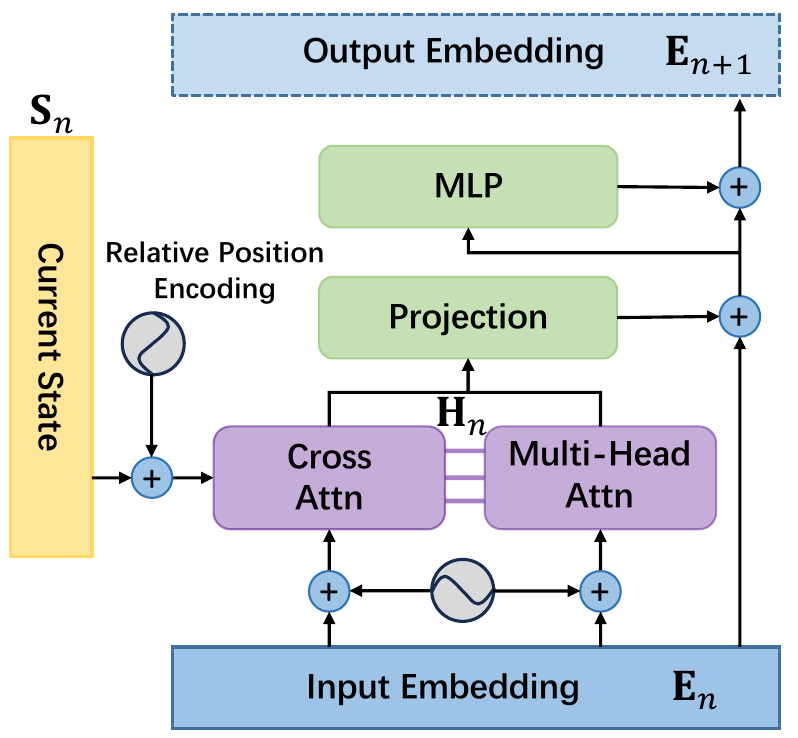}}
   \subfloat[Workflow of $\textbf{S}_{n+1}$]{%
    \label{fig:workflow_sn1}
      \includegraphics[width=0.47\columnwidth]{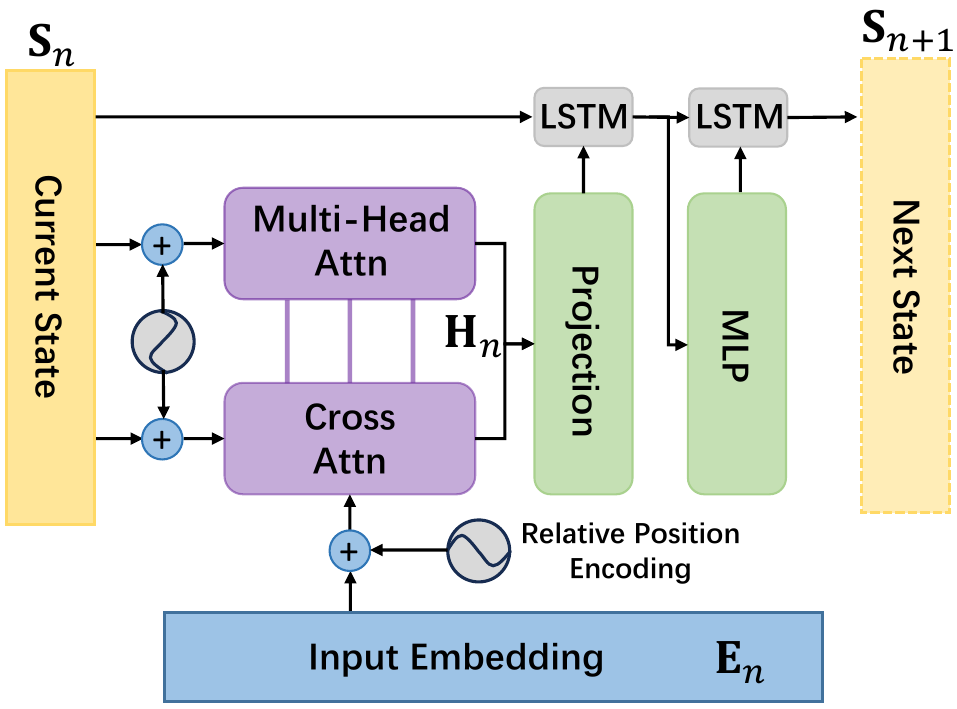}}
   \caption{Block-Recurrent Attention Layer}
   \vspace{-5mm}
\end{figure}

\subsubsection{Block-Recurrent Attention Module}
\label{br-attention}
Our model borrows the key framework of the block-recurrent transformer~\cite{hutchins2022block} to leverage the advantages of both RNN and self-attention mechanisms.
Furthermore, we incorporate relative position encoding~\cite{wu2021rethinking} as an additional component to augment the input features of the block-recurrent transformer.
This encoding improves the representation of temporal order in the model.
The block-recurrent attention module follows a pre-norm residual unit in a sandwich format.
Specifically, the merged hidden features $\textbf{M}_n$ from the merge layer are first normalized using a layernorm operation.
The normalized hidden features, or the input embedding for the next layer, can be expressed as follows: $\textbf{E}_n={\rm{LayerNorm}}(\textbf{M}_n)$.

Then the block-recurrent attention layer receives two inputs: the input embedding $\textbf{E}_n \in \mathbb{R}^{L \times D_h}$ and the current state $\textbf{S}_n \in \mathbb{R}^{L \times D_h}$.
The input embedding captures the hidden information of the IMU data in the current window segmentation, while the current state embeds the hidden information of all previous sequences. 
To generate the output embedding $\textbf{E}_{n+1}$ and the next state $\textbf{S}_{n+1}$, two distinct steps are performed, which is shown in \fig\ref{fig:workflow_en1} and \fig\ref{fig:workflow_sn1}.
To obtain $\textbf{E}_{n+1}$, the cross-attention result of the two inputs, and the multi-head self-attention~\cite{vaswani2017attention} result of the input embedding are first concatenated, which can be expressed as:
\begin{equation}
\begin{aligned}
    \textbf{H}_n={\rm{Cat}}(&{\rm{CrossAttn}}({\rm{RPE}}(\textbf{E}_n),{\rm{RPE}}(\textbf{S}_n)), \\
    &{\rm{MultiAttn}}({\rm{RPE}}(\textbf{E}_n)))
\end{aligned}
\end{equation}
where $\rm{RPE}$ denotes the relative position encoding~\cite{wu2021rethinking}.
After passing through a projection layer, $\textbf{H}_n$ is merged with $\textbf{E}_n$, and the output embedding is finally obtained through a multilayer perceptron (MLP) layer in a residue manner:
\begin{equation}
    \textbf{E}_{n+1}={\rm{Res}}({{\rm{MLP}}({{\rm{Proj}}(\textbf{H}_n})+\textbf{E}_n)})
\end{equation}

\fig\ref{fig:workflow_sn1} gives the workflow for the next state $\textbf{H}_{n+1}$.
The only difference is that it utilizes LSTM gate.
In addition, RNN structure allows the state tensor to selectively merge the current and previous information, providing essential information for future prediction.
Finally, $\textbf{E}_{n+1}$ is sent to the next module after the dropout layer.

\subsection{Multi-Loss Learning}
\label{sec:loss}
To improve \modulename's regression performance for estimating moving velocities and heading orientations, we employ a multi-loss learning strategy.
The multi-loss function consists of three components: velocity loss $\mathcal{L}_\mathbf{v}$, position loss $\mathcal{L}_\mathbf{p}$, and orientation loss $\mathcal{L}_\mathbf{o}$.

\head{Velocity Loss ($\mathcal{L}_\mathbf{v}$)}
With a predicted velocity sequence $\textbf{V}_n=[\mathbf{v}_1,\mathbf{v}_2,...,\mathbf{v}_L]$, a mean square loss is utilized to measure the difference between the ground truth velocity $\textbf{V}_n^g=[\mathbf{v}_1^g,\mathbf{v}_2^g,...,\mathbf{v}_L^g]$.
The velocity loss function can be written as $\mathcal{L}_\mathbf{v} = \sqrt{\frac{1}{L}\sum_{i=1}^L (\mathbf{v}_i-\mathbf{v}_i^g)^2}$.

\head{Position Loss ($\mathcal{L}_\mathbf{p}$)}
The predicted velocity is integrated to obtain the predicted moving trajectory $\textbf{P}_n$.
By using $\mathcal{L}_\mathbf{p}$, we aim to further reduce the cumulative error based on the learned velocity characteristics.
The equation for $\mathcal{L}_\mathbf{p}$ is $\mathcal{L}_\mathbf{p} = \sqrt{\frac{1}{L}\sum_{i=1}^L (\mathbf{p}_i-\mathbf{p}_i^g)^2}$.

\head{Orientation Loss ($\mathcal{L}_\mathbf{o}$)} 
The major prediction errors come from direction deviations, especially at low speeds.
For instance, even if the velocity loss is comparable to that at high speeds, the actual orientation error can be substantial due to the division of velocity error by the velocity modulus.
To address this issue, we incorporate orientation loss to improve the accuracy of the predicted velocity in the travel direction.
The definition of $\mathcal{L}_\mathbf{o} = \sqrt{\frac{1}{L}\sum_{i=1}^L (\frac{\mathbf{v}_i}{||\mathbf{v}_i||_2}-\frac{\mathbf{v}_i^g}{||\mathbf{v}_i^g||_2})^2}$.

The final multi-loss function is a linear combination of the above three loss functions.
In order to better balance the different loss functions and optimize the final results, we utilize the variation coefficient~\cite{groenendijk2021multi} and define the multi-loss function as:
\begin{equation}
\label{eqn:multi-loss}
    \mathcal{L}_{total} = \frac{\sigma_{\mathcal{L}_\mathbf{v}}}{\mu_{\mathcal{L}_\mathbf{v}}} \mathcal{L}_\mathbf{v} + \frac{\sigma_{\mathcal{L}_\mathbf{p}}}{\mu_{\mathcal{L}_\mathbf{p}}} \mathcal{L}_\mathbf{p} + \frac{\sigma_{\mathcal{L}_\mathbf{o}}}{\mu_{\mathcal{L}_\mathbf{o}}} \mathcal{L}_\mathbf{o}
\end{equation}
where $\mu$ and $\sigma$ denote the mean and standard deviation of the respective loss function.
The reason for applying both velocity loss and position loss is that $\mathcal{L}_\mathbf{v}$ corrects the velocity vector on each frame, while $\mathcal{L}_\mathbf{p}$ corrects it on a time window scale.
By combining local and global information, the model can learn the characteristics of the motion better.
A detailed ablation analysis can be found in \S\ref{sec:abalation-study}.

\subsection{Data Augmentation}
\label{sec:data-augmentation}
To improve the robustness of \modulename under different heading directions, both IMU data $\textbf{X}_n$ and ground truth velocity $\textbf{V}_n^g$ are randomly rotated by an angle of  $\phi \in [0,2\pi)$ on the floor plane~\cite{cao2022rio}:
\begin{equation}
\label{eqn:random-rotation}
{\Delta\textbf{q}}\textbf{V}_n^g{\Delta\textbf{q}^{*}}={\Delta\textbf{q}}\textbf{X}_n{\Delta\textbf{q}^{*}},
\end{equation}
where ${\Delta\textbf{q}}$ represents the random rotation in quaternion.
This data augmentation helps distribute movement orientations more evenly on the floor plane, improving model robustness.
An ablation study in \S\ref{sec:aug-study} further supports its effectiveness.

\begin{figure}[t]
\centering
   \subfloat[System Setup]{%
    \label{fig:system-set}
      \includegraphics[width=0.3\columnwidth]{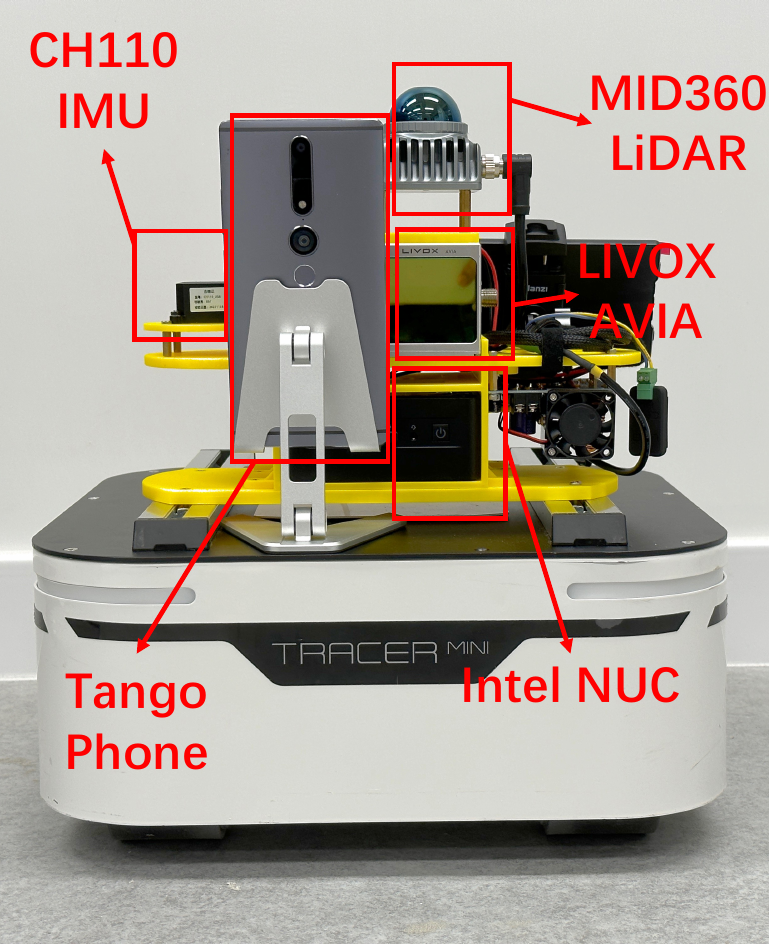}
	  }%
   \subfloat[Building Layouts]{%
    \label{fig:buildings}
      \includegraphics[width=0.52\columnwidth]{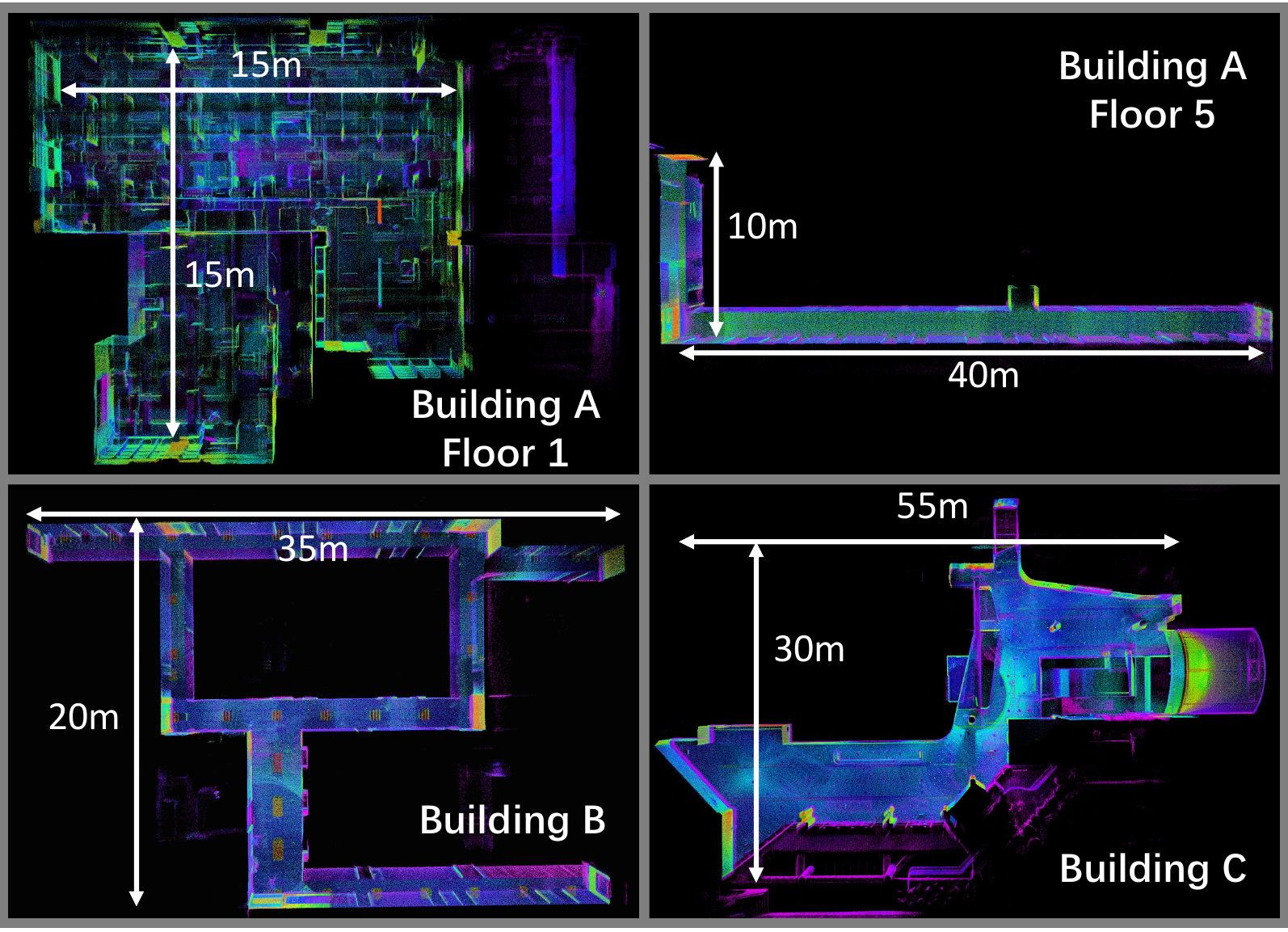}
	  }%
   \caption{Experimental Settings. \rm{The floorplans are visualized based on the results reconstructed by FAST-LIO.}}
   \vspace{-2mm}
\end{figure}

\section{Implementation}
\label{sec:implementation}
We build a customized robotic platform for data collection, which captures raw IMU data as \sysname input, visual-inertial tracking results from a commercialized device (Tango Phone
\footnote{\url{https://www.lenovo.com/il/en/tango/}}
) for comparison, and the ground truth data from a LiDAR-inertial tracking system implemented with FAST-LIO~\cite{xu2021fast}.
We choose the Tango phone and FAST-LIO for specific reasons.
Tango phone has been used as the ground truth in inertial tracking datasets like RIDI and RONIN.
FAST-LIO is known for its minimal drift in indoor environments.

\fig\ref{fig:system-set} illustrates the system setup.
The core of the system is an Intel NUC with an i7-1260P CPU.
It runs the FAST-LIO algorithm using point clouds from the LIVOX MID360\footnote{FOV: 360°×59°; Point Cloud Density: 40-line. \url{https://www.livoxtech.com/mid-360}} and its built-in IMU to determine the robot’s indoor position.
A HiPNUC CH110 9-axis IMU is mounted on the left side to collect raw data for model input.
A Tango phone, placed at the front, provides visual-inertial tracking results for comparison in \S\ref{sec:vslam-compare}.
In addition, we add AVIA
\footnote{FOV: 70.4°*77.2°; \url{https://www.livoxtech.com/avia}}
, a narrow-FOV LiDAR, to demonstrate the fact that LiDAR-inertial tracking systems cannot work well in plain indoor environments.
The robot's territory is AgileX Tracer-Mini, 
whose size is $40\rm{cm}\times40\rm{cm}\times26\rm{cm}$ with a maximum speed of 1.5m/s.

\begin{table}[t]
\centering
\caption{Summary of datasets}
\label{tab:datasets}
\resizebox{0.95\columnwidth}{!}{%
\begin{tabular}{cccccc}
\hline
Dataset & Year & IMU Carrier           & \begin{tabular}[c]{@{}c@{}}Sample\\ Frequency\end{tabular} & \begin{tabular}[c]{@{}c@{}}\# of\\ Sequences\end{tabular} & Ground Truth       \\ \hline
NeurIT  & 2023 & CH110                 & 200 Hz                                                     & 110                                                       & FAST-LIO           \\ \hline
RIDI    & 2017 & Lenovo Phab2 Pro      & 200 Hz                                                     & 98                                                        & Google Tango phone \\ \hline
RONIN   & 2019 & Galaxy S9, Pixel 2 XL & 200 Hz                                                     & 118                                                       & Asus Zenfone AR    \\ \hline
\end{tabular}%
}
\end{table}

\begin{figure}[t]
  \begin{center}
  \includegraphics[width=0.4\textwidth]{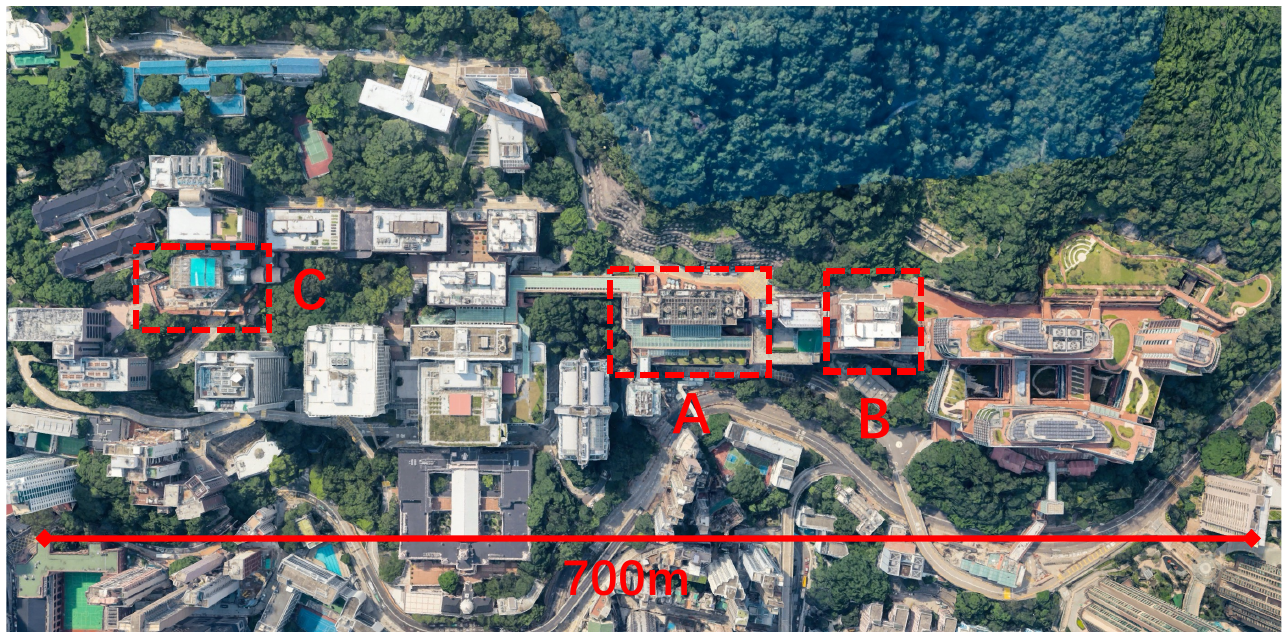}
  \caption{Topview of the locations of three buildings.}\label{fig:top-view}
  \end{center}
  \vspace{-5mm}
\end{figure}

\section{Experiments}
\label{sec:exp}
We evaluate \sysname across diverse datasets and metrics.
To test its robustness, we conduct experiments in large urban spaces, such as shopping malls and concert halls.
We also compare it with a visual-inertial method indoors, perform ablation studies to analyze component contributions, and assess real-time performance to validate practical deployment.

\subsection{Datasets}
\label{dataset-baseline}
Due to the lack of IMU datasets on robot subjects, we built our dataset to verify the effectiveness of \sysname.
We split the dataset into four parts, \ie, training, validation, test-seen, and test-unseen sets.
More details are described below:

\head{\sysname Dataset}
We built the \sysname dataset using the customized robotic platform on four floors across three buildings, which feature different layouts, environments, and floor conditions, as shown in \fig\ref{fig:buildings}.
The locations of these three buildings are illustrated in \fig\ref{fig:top-view}.
We collect the training, validation, and test-seen sets in Building A, and build the test-unseen set in Building B and C.
During data collection, the robot moves at varying speeds up to the maximum value (1.5m/s). 

The dataset contains 110 sequences, totaling around 15 hours of tracking data that corresponds to a travel distance of around 33.7 km.
Each sequence of data lasts 6$\sim$10 minutes. 
The ratio of the training set, validation set, test-seen set, and test-unseen set is 15:3:3:4.

\head{RIDI}
RIDI contains the walking data of 10 human participants.
We follow the default split setting in this dataset, which is 5:1:2:1. 

\head{RONIN}
RONIN involves 100 human subjects. 
It is divided according to the default setting, \ie, 69:16:35:32. 
It should be noted that only 50\% of the dataset has been publicly released.

During the training stage, we divide the \sysname and RIDI according to the segmentation size of $200\times15$ and the step size of $400$, and slide $15$ times according to the window size of $200$ on each segment sequence.
We partition the RONIN dataset into segments of size $400 \times 20$ with a step size of 1000. 
Each segment is then processed using a sliding window of size 400, applied 20 times per sequence.
During testing, we use a window size and step size of 200 for both \sysname and RIDI, and 400 for RONIN.. 

\vspace{-5mm}

\subsection{Baselines}
To comprehensively evaluate \sysname, we compare it with seven state-of-the-art baseline approaches.

\begin{itemize}[leftmargin=*]
    \item \head{Naive Double Integration (NDI)} 
    NDI~\cite{savage1998strapdown} calculates current motion by integrating gyroscopes and accelerations.
    \item \head{Extended Kalman Filter (EKF)} EKF~\cite{ribeiro2004kalman} filters out Gaussian-like noise in the IMU data and combines it with the magnetometer to correct the orientation error. We use the python implementation~\cite{aipiano2025esekf_imu} in this work.
    \item \head{IONet} IONet~\cite{chen2018ionet} employs RNN and fully-connected layers for model design.
    It only takes accelerations and gyroscopes as input features.
    \item \head{RONIN (Bi-LSTM)} 
    We adopt the Bi-LSTM model that reports the best performance in RONIN~\cite{yan2019ronin} dataset.
    Only accelerations and gyroscopes are taken as input in RONIN.
    \item \head{LLIO}
    LLIO~\cite{wang2022llio} utilizes multiple MLP layers in a residual way to predict the trajectories for pedestrians.
    The input features consist of accelerations and gyroscopes.
    \item \head{UniTS}
    UniTS \cite{li2021units} is an STFT-inspired neural network for sensory time series classification.
    UniTS takes accelerations and gyroscopes as inputs, the same as IONet and RONIN.
    \item \head{RF-Net} 
    RF-Net~\cite{ding2020rf} is designed for wireless sensing, which learns from time-frequency representations.
    Only accelerations and gyroscopes are taken as input features.
\end{itemize}

It should be noted that Transformer-based solutions, such as CTIN~\cite{rao2022ctin} and A2DIO~\cite{wang2022a2dio}, even though they achieve promising results in inertial tracking, the model cannot be reproduced without implementation details and the source code.
In this case, we do not include them as baseline models.
We will evaluate the performance comparison with Transformer-based methodology in the ablation study \S\ref{sec:abalation-study}. 

\begin{table*}[t]
\centering
\caption{Evaluation on \sysname Dataset: {\rm (\textcolor[HTML]{FF0000}{{\textbf{Red}}}: the best performance among all algorithms. \textcolor[HTML]{00B0F0}{\textbf{Blue}}: the best performance among all baseline models. Improvement is calculated based on the best baseline.)}}
\label{tab:neurit-eval}
\resizebox{0.85\textwidth}{!}{%
\begin{tabular}{c|c|c|c|c|c|c|c|c|c|cc|cc}
\hline
\multicolumn{1}{l|}{Dataset} &
  \begin{tabular}[c]{@{}c@{}}Test\\ Subject\end{tabular} &
  Metric &
  NDI &
  EKF &
  RONIN &
  IONet &
  LLIO &
  UniTS &
  RF-Net &
  \begin{tabular}[c]{@{}c@{}}NeurIT\\ w/o magn\end{tabular} &
  Impvt &
  NeurIT &
  Impvt \\ \hline
 &
   &
  ATE &
  7.926 &
  7.651 &
  3.590 &
  3.262 &
  {\color[HTML]{333333} 3.061} &
  {\color[HTML]{00B0F0} \textbf{2.712}} &
  5.353 &
  \textbf{1.539} &
  \textbf{43.26\% ↑} &
  {\color[HTML]{FF0000} \textbf{1.010}} &
  \textbf{62.76\% ↑} \\ \cline{3-14} 
 &
   &
  RTE &
  10.005 &
  9.608 &
  2.190 &
  1.584 &
  {\color[HTML]{333333} 1.592} &
  {\color[HTML]{00B0F0} \textbf{1.370}} &
  2.085 &
  \textbf{1.083} &
  \textbf{20.95\% ↑} &
  {\color[HTML]{FF0000} \textbf{0.580}} &
  \textbf{57.67\% ↑} \\ \cline{3-14} 
 &
   &
  PDE &
  0.029 &
  0.040 &
  0.026 &
  0.027 &
  {\color[HTML]{333333} 0.024} &
  {\color[HTML]{00B0F0} \textbf{0.021}} &
  0.048 &
  \textbf{0.010} &
  \textbf{51.31\% ↑} &
  {\color[HTML]{FF0000} \textbf{0.007}} &
  \textbf{67.38\% ↑} \\ \cline{3-14} 
 &
  \multirow{-4}{*}{Seen} &
  AYE &
  123.086 &
  124.774 &
  63.770 &
  {\color[HTML]{00B0F0} \textbf{46.228}} &
  58.618 &
  121.414 &
  99.765 &
  \textbf{27.544} &
  \textbf{40.42\% ↑} &
  {\color[HTML]{FF0000} \textbf{35.990}} &
  \textbf{22.15\% ↑} \\ \cline{2-14} 
 &
   &
  ATE &
  9.238 &
  8.958 &
  4.260 &
  3.446 &
  {\color[HTML]{333333} 3.167} &
  {\color[HTML]{00B0F0} \textbf{2.904}} &
  4.125 &
  \textbf{1.755} &
  \textbf{39.56\% ↑} &
  {\color[HTML]{FF0000} \textbf{1.140}} &
  \textbf{60.75\% ↑} \\ \cline{3-14} 
 &
   &
  RTE &
  11.241 &
  11.099 &
  2.420 &
  1.874 &
  1.894 &
  {\color[HTML]{00B0F0} \textbf{1.736}} &
  2.236 &
  \textbf{1.538} &
  \textbf{11.38\% ↑} &
  {\color[HTML]{FF0000} \textbf{0.760}} &
  \textbf{56.21\% ↑} \\ \cline{3-14} 
 &
   &
  PDE &
  {\color[HTML]{00B0F0} \textbf{0.020}} &
  0.036 &
  0.035 &
  0.027 &
  0.024 &
  0.023 &
  0.035 &
  \textbf{0.013} &
  \textbf{32.88\% ↑} &
  {\color[HTML]{FF0000} \textbf{0.009}} &
  \textbf{55.01\% ↑} \\ \cline{3-14} 
\multirow{-8}{*}{NeurIT} &
  \multirow{-4}{*}{Unseen} &
  AYE &
  132.905 &
  134.107 &
  51.660 &
  {\color[HTML]{00B0F0} \textbf{40.101}} &
  60.501 &
  119.548 &
  89.364 &
  \textbf{22.950} &
  \textbf{42.77\% ↑} &
  {\color[HTML]{FF0000} \textbf{32.140}} &
  \textbf{19.86\% ↑} \\ \hline
\end{tabular}%
}
\end{table*}

\begin{table*}[ht]
\centering
\caption{Evaluation on RIDI \& RONIN Datasets: {\rm (\textcolor[HTML]{FF0000}{{\textbf{Red}}}: the best performance among all algorithms. \textcolor[HTML]{00B0F0}{\textbf{Blue}}: the best performance among all baseline models. Improvement is calculated based on the best baseline.)}}
\label{tab:ridi-ronin}
\resizebox{0.85\textwidth}{!}{%
\begin{tabular}{c|c|c|c|c|c|c|c|c|c|cc|cc}
\hline
Dataset &
  \begin{tabular}[c]{@{}c@{}}Test\\ Subject\end{tabular} &
  Metric &
  NDI &
  EKF &
  RONIN &
  IONet &
  LLIO &
  UniTS &
  RF-Net &
  \begin{tabular}[c]{@{}c@{}}NeurIT\\ w/o magn\end{tabular} &
  Impvt &
  NeurIT &
  Impvt \\ \hline
 &
   &
  ATE &
  16.75 &
  26.04 &
  1.72 &
  1.76 &
  1.72 &
  {\color[HTML]{00B0F0} \textbf{1.66}} &
  3.94 &
  \textbf{1.63} &
  \textbf{1.40\% ↑} &
  {\color[HTML]{FF0000} \textbf{1.53}} &
  \textbf{7.64\%↑} \\ \cline{3-14} 
 &
   &
  RTE &
  17.58 &
  37.51 &
  {\color[HTML]{00B0F0} \textbf{2.04}} &
  2.09 &
  2.17 &
  2.16 &
  4.34 &
  \textbf{2.04} &
  \textbf{0.16\% ↑} &
  {\color[HTML]{FF0000} \textbf{2.02}} &
  \textbf{1.19\% ↑} \\ \cline{3-14} 
 &
   &
  PDE &
  0.203 &
  0.684 &
  0.0328 &
  {\color[HTML]{00B0F0} \textbf{0.0319}} &
  0.0345 &
  0.0338 &
  0.0688 &
  \textbf{0.0317} &
  \textbf{0.55\% ↑} &
  {\color[HTML]{FF0000} \textbf{0.0286}} &
  \textbf{10.41\% ↑} \\ \cline{3-14} 
 &
  \multirow{-4}{*}{Seen} &
  AYE &
  127.97 &
  114.15 &
  {\color[HTML]{00B0F0} \textbf{51.39}} &
  57.52 &
  54.71 &
  70.95 &
  63.74 &
  \textbf{27.21} &
  \textbf{47.05\% ↑} &
  {\color[HTML]{FF0000} \textbf{26.33}} &
  \textbf{48.76\% ↑} \\ \cline{2-14} 
 &
   &
  ATE &
  17.30 &
  24.14 &
  {\color[HTML]{00B0F0} \textbf{1.87}} &
  2.10 &
  1.96 &
  1.99 &
  3.20 &
  \textbf{1.51} &
  \textbf{19.10\% ↑} &
  {\color[HTML]{FF0000} \textbf{1.51}} &
  \textbf{19.10\% ↑} \\ \cline{3-14} 
 &
   &
  RTE &
  18.02 &
  40.67 &
  1.87 &
  2.18 &
  {\color[HTML]{00B0F0} \textbf{1.87}} &
  1.92 &
  3.20 &
  \textbf{1.66} &
  \textbf{11.10\% ↑} &
  {\color[HTML]{FF0000} \textbf{1.60}} &
  \textbf{14.31\% ↑} \\ \cline{3-14} 
 &
   &
  PDE &
  0.1799 &
  0.7095 &
  {\color[HTML]{00B0F0} \textbf{0.0291}} &
  0.0354 &
  0.0297 &
  0.0310 &
  0.0577 &
  \textbf{0.0278} &
  \textbf{4.62\% ↑} &
  {\color[HTML]{FF0000} \textbf{0.0243}} &
  \textbf{16.55\% ↑} \\ \cline{3-14} 
\multirow{-8}{*}{RIDI} &
  \multirow{-4}{*}{Unseen} &
  AYE &
  130.59 &
  105.93 &
  {\color[HTML]{00B0F0} \textbf{48.47}} &
  53.67 &
  51.82 &
  67.42 &
  64.64 &
  \textbf{24.34} &
  \textbf{49.78\% ↑} &
  {\color[HTML]{FF0000} \textbf{24.21}} &
  \textbf{50.05\% ↑} \\ \hline
 &
   &
  ATE &
  22.86 &
  33.26 &
  7.76 &
  6.20 &
  6.18 &
  {\color[HTML]{00B0F0} \textbf{5.68}} &
  8.78 &
  \textbf{4.58} &
  \textbf{19.38\% ↑} &
  {\color[HTML]{FF0000} \textbf{4.37}} &
  \textbf{23.10\% ↑} \\ \cline{3-14} 
 &
   &
  RTE &
  30.49 &
  35.10 &
  3.19 &
  3.17 &
  3.04 &
  {\color[HTML]{00B0F0} \textbf{2.77}} &
  6.43 &
  \textbf{2.60} &
  \textbf{6.03\% ↑} &
  {\color[HTML]{FF0000} \textbf{2.59}} &
  \textbf{6.53\% ↑} \\ \cline{3-14} 
 &
   &
  PDE &
  0.124 &
  0.107 &
  0.0432 &
  0.0390 &
  0.0312 &
  {\color[HTML]{00B0F0} \textbf{0.0267}} &
  0.0493 &
  \textbf{0.0216} &
  \textbf{18.99\% ↑} &
  {\color[HTML]{FF0000} \textbf{0.0181}} &
  \textbf{31.99\% ↑} \\ \cline{3-14} 
 &
  \multirow{-4}{*}{Seen} &
  AYE &
  122.48 &
  120.96 &
  {\color[HTML]{00B0F0} \textbf{63.12}} &
  74.49 &
  69.059 &
  107.92 &
  86.91 &
  {\color[HTML]{FF0000} \textbf{46.56}} &
  \textbf{26.24\% ↑} &
  \textbf{47.60} &
  \textbf{24.59\% ↑} \\ \cline{2-14} 
 &
   &
  ATE &
  17.30 &
  33.47 &
  8.91 &
  7.73 &
  7.57 &
  {\color[HTML]{00B0F0} \textbf{6.38}} &
  10.27 &
  \textbf{5.78} &
  \textbf{9.52\% ↑} &
  {\color[HTML]{FF0000} \textbf{5.35}} &
  \textbf{16.26\% ↑} \\ \cline{3-14} 
 &
   &
  RTE &
  18.02 &
  41.07 &
  4.12 &
  4.63 &
  {\color[HTML]{00B0F0} \textbf{3.85}} &
  3.86 &
  8.20 &
  \textbf{3.59} &
  \textbf{6.85\% ↑} &
  {\color[HTML]{FF0000} \textbf{3.55}} &
  \textbf{7.89\% ↑} \\ \cline{3-14} 
 &
   &
  PDE &
  0.0912 &
  0.3166 &
  0.0462 &
  0.0358 &
  0.03622 &
  {\color[HTML]{00B0F0} \textbf{0.0320}} &
  0.0592 &
  \textbf{0.0268} &
  \textbf{16.25\% ↑} &
  {\color[HTML]{FF0000} \textbf{0.0243}} &
  \textbf{24.06\% ↑} \\ \cline{3-14} 
\multirow{-8}{*}{RONIN} &
  \multirow{-4}{*}{Unseen} &
  AYE &
  110.00 &
  114.80 &
  {\color[HTML]{00B0F0} \textbf{64.46}} &
  76.49 &
  69.21 &
  106.80 &
  88.69 &
  \textbf{47.88} &
  \textbf{25.72\% ↑} &
  {\color[HTML]{FF0000} \textbf{46.66}} &
  \textbf{27.61\% ↑} \\ \hline
\end{tabular}%
}
\vspace{-3mm}
\end{table*}

\subsection{Evaluation Metrics}
\label{evaluation-metrics}
The evaluation mainly focuses on the predicted locations $\mathbf{P}$ derived from $\mathbf{V}$.
Particularly, we consider four common criteria to assess the error between the estimated positions $\mathbf{P}$ and the ground truths $\mathbf{P}^g$.

\begin{itemize}[leftmargin=*]
    \item \textbf{Absolute Trajectory Error (ATE)}: ATE~\cite{sturm2012benchmark} calculates the root mean square error (RMSE) of the predicted trajectory and the ground truth trajectory, which is given as $\sqrt{\frac{1}{n}\sum_i\Vert \mathbf{p}_i-\mathbf{p}^g_i\Vert_2^2}$.
    \item \textbf{Relative Trajectory Error (RTE)}: RTE~\cite{sturm2012benchmark} measures the RMSE between the predicted trajectory and the ground truth over a time interval. RTE can be calculated as $\sqrt{\frac{1}{n}\sum_i\left\Vert(\mathbf{p}_{i+\Delta t}-\mathbf{p}_i)-(\mathbf{p}^g_{i+\Delta t}-\mathbf{p}^g_i)\right\Vert_2^2}$, where $\Delta t$ is set to 60s in our evaluation.
    \item \textbf{Position Drift Error (PDE)}: PDE~\cite{rao2022ctin} computes the final position drift over the whole sequence. It can be expressed as $\Vert \mathbf{p}_N-\mathbf{p}^g_N\Vert_2/{D}$, where $N$ is the last index of the sequence, and $D$ denotes the ground truth trajectory length.
    \item \textbf{Absolute Yaw Error (AYE)}: AYE~\cite{wang2022llio} measures the RMSE between the heading orientations between ground truth and predicted trajectories, \ie, $\sqrt{\frac{1}{n}\sum_i\left\Vert\frac{\mathbf{v}_i}{\Vert \mathbf{v}_i\Vert_2}-\frac{\mathbf{v}^g_i}{\Vert \mathbf{v}^g_i\Vert_2}\right\Vert_2^2}$.
\end{itemize}

\begin{figure*}[t]
  \begin{center}
  \includegraphics[width=0.85\textwidth]{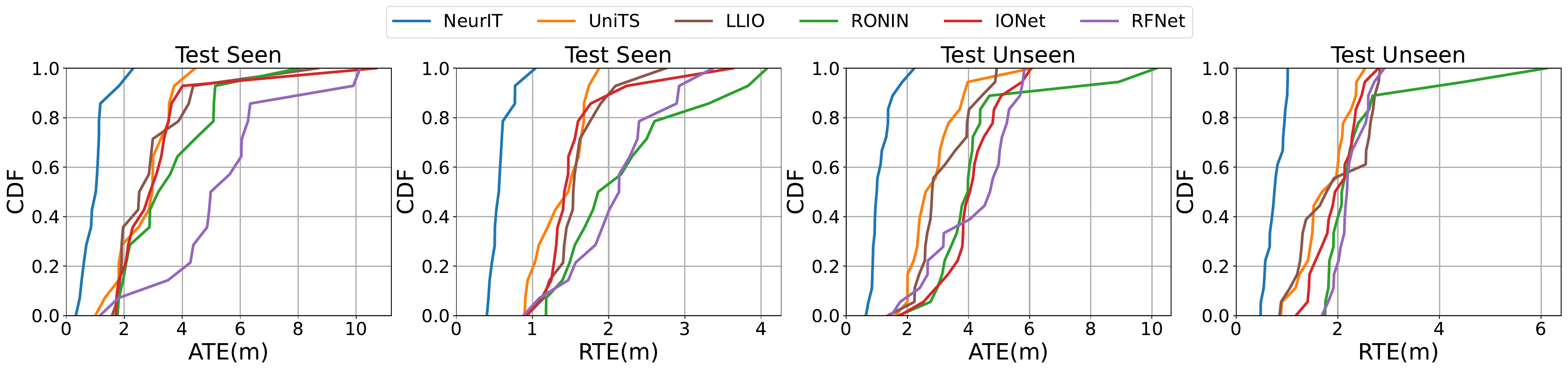}
  \caption{CDF of ATE and RTE on \sysname Test Dataset}\label{fig:cdf}
  \end{center}
  \vspace{-5mm}
\end{figure*}

\begin{figure*}[t]
  \begin{center}
  \includegraphics[width=0.85\textwidth]{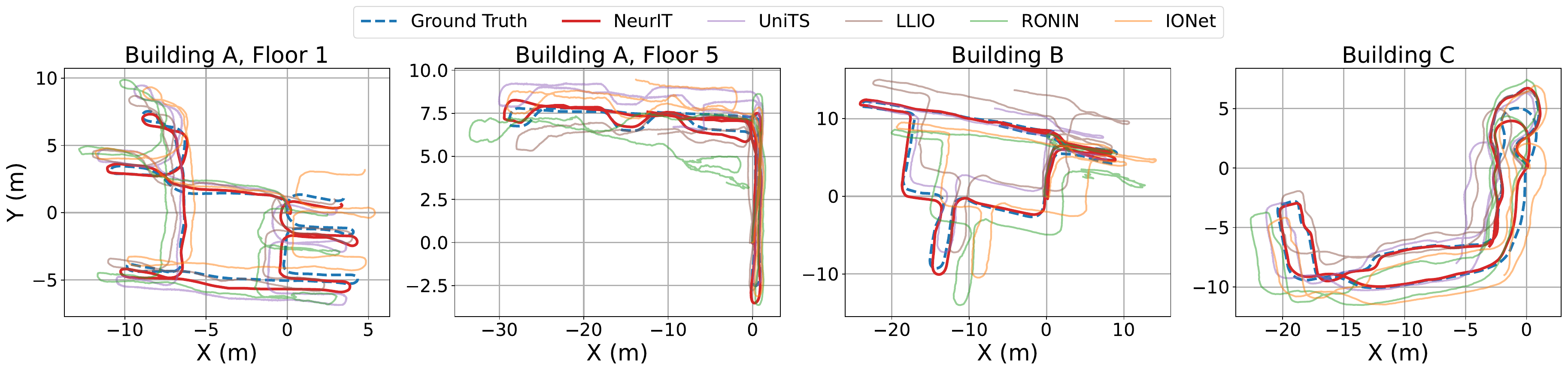}
  \caption{Examples of Predicted Trajectories of \sysname}\label{fig:traj-visualization}
  \end{center}
  \vspace{-8mm}
\end{figure*}

\section{Evaluation}
\label{sec:evaluation}

\subsection{Overall Performance}
\label{neurit-performance}

\head{Evaluation on \sysname Dataset} 
\tab\ref{tab:neurit-eval} summarizes the overall evaluation on the \sysname dataset.
\sysname achieves outstanding accuracy on both seen and unseen test sets, with a PDE of 0.08, equivalent to just 0.8cm per meter traveled.
It reduces the average ATE by 62\% compared to the best baseline, and maintains a minimal RTE of 0.6\%, corresponding to 0.6m drift over 1 minute, whereas UniTS reports over double that error ($\sim$1.5m).
LLIO ranks second among baselines, but shows a performance gap likely due to less effective feature design.
Thanks to multi-loss learning, \sysname also improves AYE, reducing heading error from 43° (IONet) to 33°, demonstrating better orientation estimation.
Even without magnetometer inputs, \sysname outperforms the best baseline by 30\% across four metrics.
When differential magnetometers are incorporated, performance improves further, particularly RTE, which is cut by nearly half, highlighting the effectiveness of our sensor fusion design.

\begin{figure}[t]
\centering
   \subfloat[ATE of different moving time.]{%
    \label{fig:ate-time}
      \includegraphics[width=0.47\columnwidth]{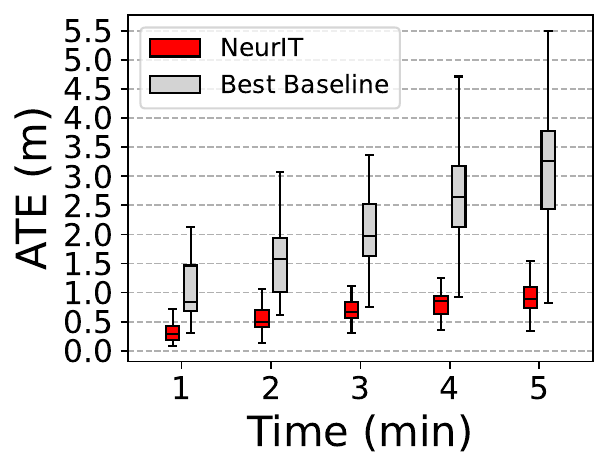}
	  }%
   \subfloat[ATE of different moving distance.]{%
    \label{fig:ate-dis}
      \includegraphics[width=0.47\columnwidth]{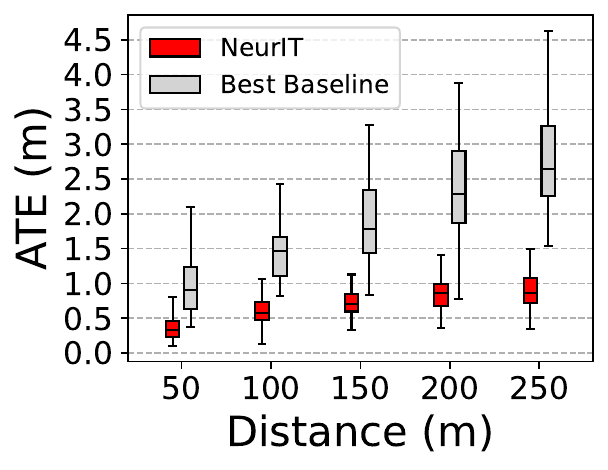}
	  }%
   \caption{Performance comparison between NEURIT and the best baseline model (UniTS) over time and distance.}
   \vspace{-5mm}
\end{figure}

\begin{figure}[t]
\centering
   \subfloat[]{%
    \label{fig:vec-loss}
      \includegraphics[width=0.47\columnwidth]{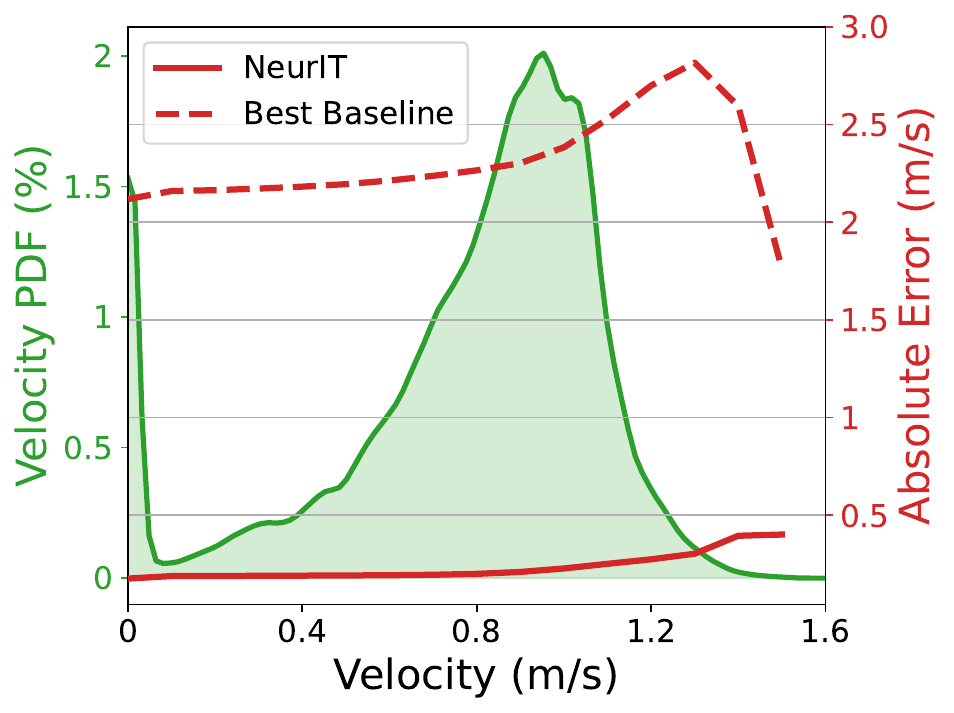}
	  }%
   \subfloat[]{%
    \label{fig:vec-loss-traj}
      \includegraphics[width=0.47\columnwidth]{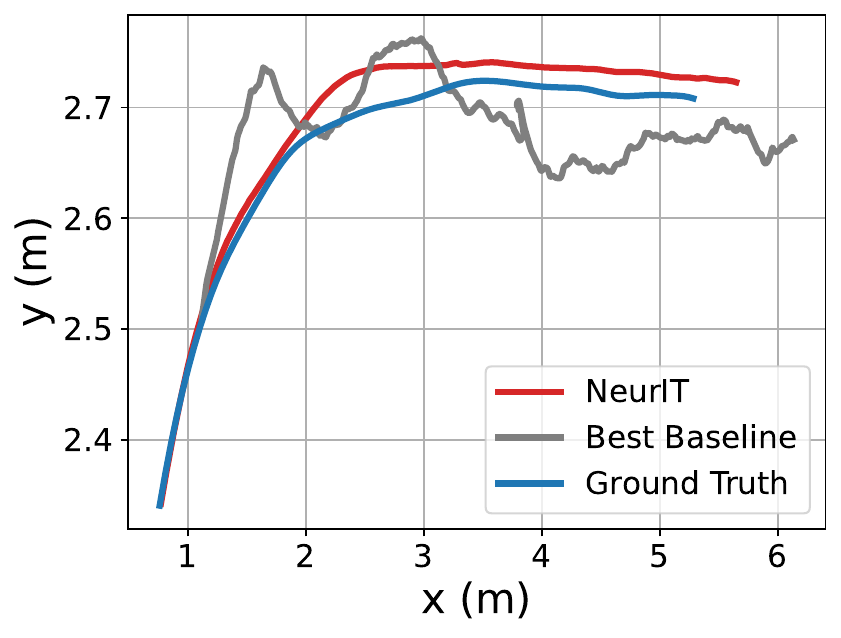}
	  }%
   \caption{Velocity prediction comparison with the best baseline (UniTS). (a): Absolute velocity error at different speeds. (b): Comparison of predicted trajectories in 5s.}
   \vspace{-5mm}
\end{figure}

\fig\ref{fig:cdf} shows the cumulative distribution of ATE and RTE, where \sysname consistently outperforms all baselines, especially in unseen environments.
Even in the worst-case scenario, \sysname exceeds the best-case performance of other models.
Trajectory visualizations in \fig\ref{fig:traj-visualization} confirm \sysname’s stable tracking, with predicted paths closely matching ground truth, unlike other methods that show significant deviations.

Additional comparisons reveal the limitations of using only RNNs or Transformers.
While UniTS performs well overall, it suffers from high AYE and occasional failure cases, leading to long-tail drift.
In contrast, \sysname effectively combines RNN and Transformer strengths, achieving state-of-the-art accuracy and robustness.

\head{Evaluation on RIDI \& RONIN Datasets} 
\label{ridi-ronin}
While this work primarily focuses on robotic tracking, we also evaluate \sysname on pedestrian tracking datasets to assess its generalization capability.
As shown in \tab\ref{tab:ridi-ronin}, \sysname outperforms all baseline models on both the RIDI and RONIN datasets.
Although baselines perform reasonably well on seen data, their accuracy drops significantly on unseen test sets, where \sysname achieves notably better results.
On RIDI, the performance gain rises from 17\% on the seen set to nearly 25\% on the unseen set.
For the RONIN dataset, only half of the total data is publicly available, limiting the model's ability to learn comprehensive pedestrian features.
This constraint leads to a smaller improvement in unseen conditions and generally weaker performance compared to RIDI.
Overall, the strong results across both domains highlight \sysname’s potential as a unified model for both robotic and pedestrian inertial tracking.

\subsection{Benchmark Study}
\label{sec:benchmark-study}

\head{Performance over Time and Distance}
\label{time-distance-study}
We analyze how prediction errors grow over time and distance.
As shown in \fig\ref{fig:ate-time} and \fig\ref{fig:ate-dis}, error increases with longer tracking durations and distances, as expected.
However, \sysname shows a significantly slower error growth compared to the best baseline, UniTS.
For instance, after 250 meters, \sysname's ATE remains lower than UniTS’s ATE at just 50 meters.
Over a 5-minute or 250-meter trajectory, \sysname maintains a mean error of about 0.9m, whereas UniTS reaches the same error within 1 minute and exceeds 2.6m after 5 minutes.
These results demonstrate the strong stability and robustness of \sysname for long-duration indoor tracking.

\begin{figure}[t]
  \begin{center}
  \includegraphics[width=0.85\columnwidth]{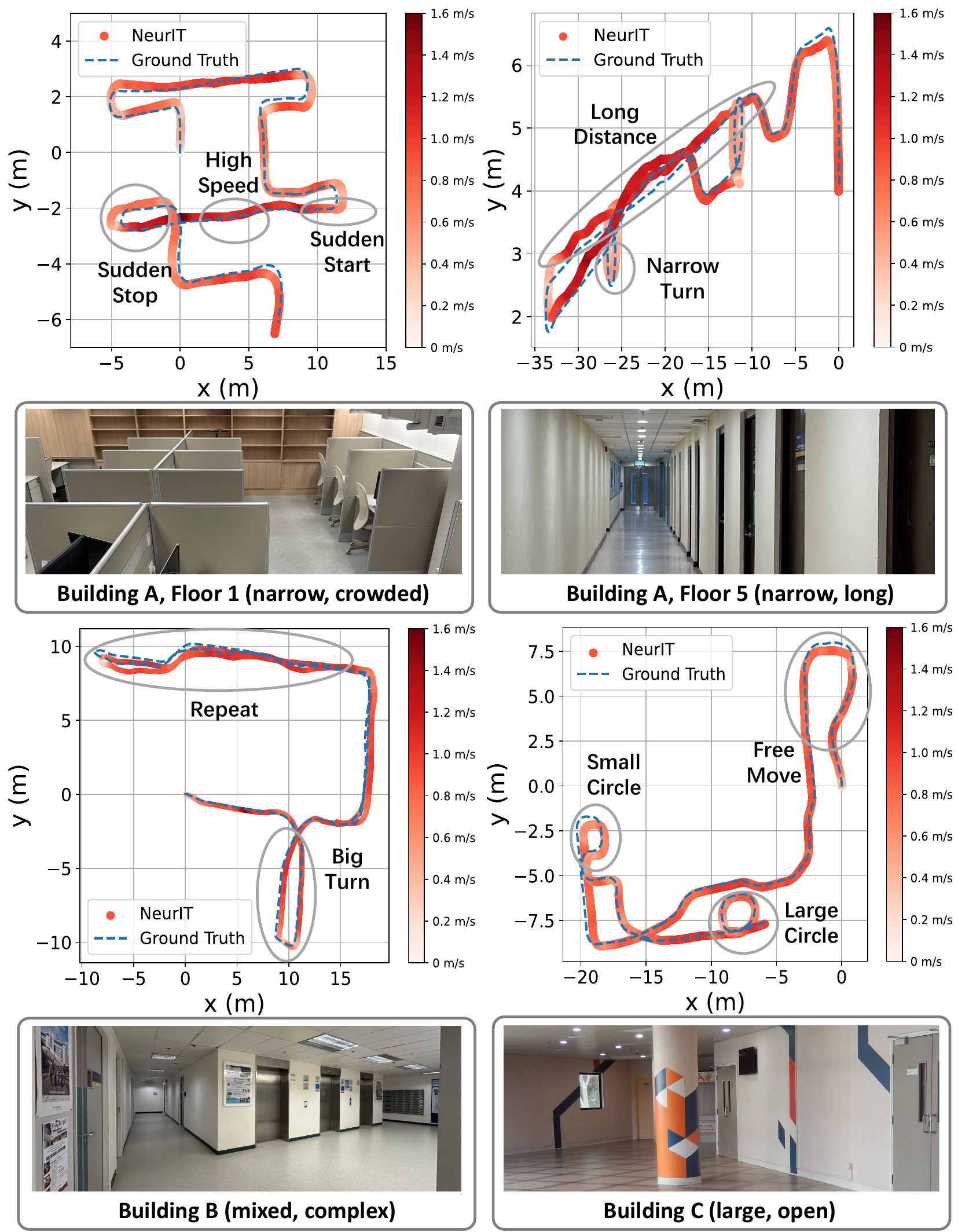}
  \caption{Visualization of robot motion status in different indoor environments.}\label{fig:vis-3b4f}
  \end{center}
  \vspace{-8mm}
\end{figure}

\head{Velocity Prediction Error}
\label{vec-loss}
To evaluate \sysname’s performance in speed estimation, we compare its predicted speed against ground truth.
As shown in \fig\ref{fig:vec-loss}, \sysname consistently outperforms the best baseline (UniTS), with an average error margin of about 0.2m/s.
The error remains low and stable, under 0.4m/s, until speeds exceed 1.2m/s, beyond which it grows slightly, likely due to limited high-speed samples in the training data.
This suggests that including more high-speed instances could further improve accuracy.
In contrast, UniTS performs poorly across all speeds, with an average error exceeding 2m/s.
As illustrated in \fig\ref{fig:vec-loss-traj}, \sysname generates a smooth and accurate trajectory that closely follows the ground truth, while the baseline produces erratic and imprecise movements.

\head{Different Motion Status}
Robotic movement often adapts to environmental conditions.
To assess \sysname’s robustness across varied indoor settings, we conduct additional experiments and visualize the predicted trajectories and corresponding speeds in \fig\ref{fig:vis-3b4f}.
In Building A, floor 1, a narrow, crowded office space, \sysname accurately tracks the robot through tight corridors, including turns and straight segments.
It also handles challenging cases like sudden starts and high-speed linear motion with ease.
Building A, floor 5, presents a more extreme scenario, featuring a 40-meter-long narrow corridor.
Even in new environments such as Buildings B and C, representing complex and open spaces, \sysname maintains high accuracy despite not having seen these conditions during training. These results demonstrate \sysname’s strong robustness and adaptability across diverse and unseen indoor environments.

\begin{figure}[t]
  \begin{center}
  \includegraphics[width=0.8\columnwidth]{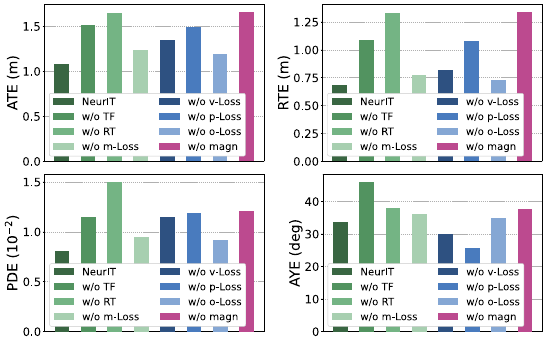}
  \caption{Ablation Study on \sysname. The green bars: Modules in \modulename; Blue bars: Loss functions; Purple bar: Magnetometers.}\label{fig:ablation}
  \end{center}
  \vspace{-5mm}
\end{figure}

\begin{figure}[t]
\centering
   \subfloat[Overall performance]{%
    \label{fig:aug-test}
      \includegraphics[width=0.3\columnwidth]{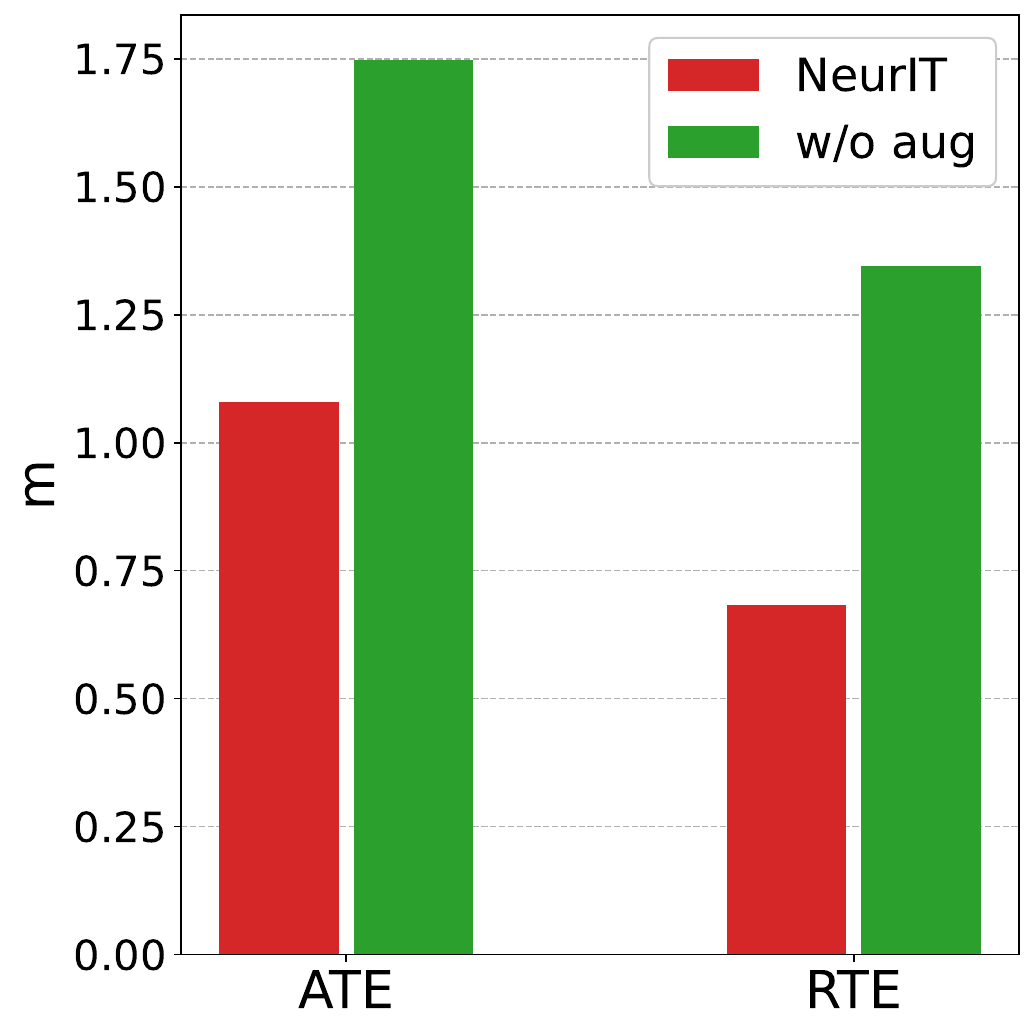}
	  }%
   \subfloat[with augmentation]{%
    \label{fig:aug-vis}
      \includegraphics[width=0.3\columnwidth]{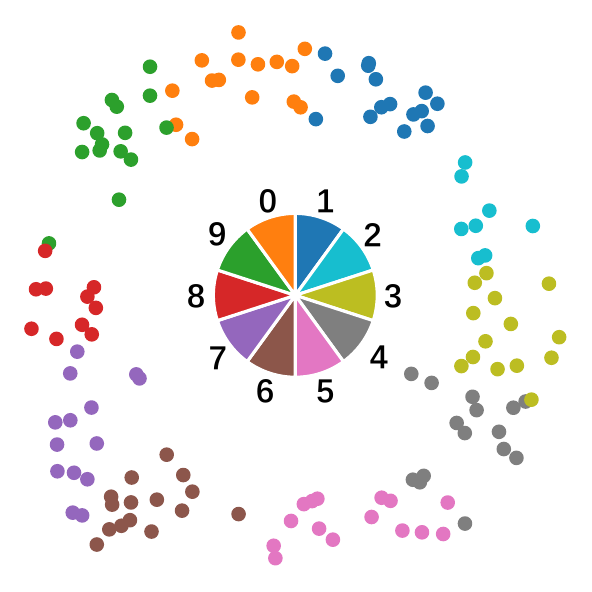}
	  }%
   \subfloat[without augmentation]{%
    \label{fig:woaug-vis}
      \includegraphics[width=0.3\columnwidth]{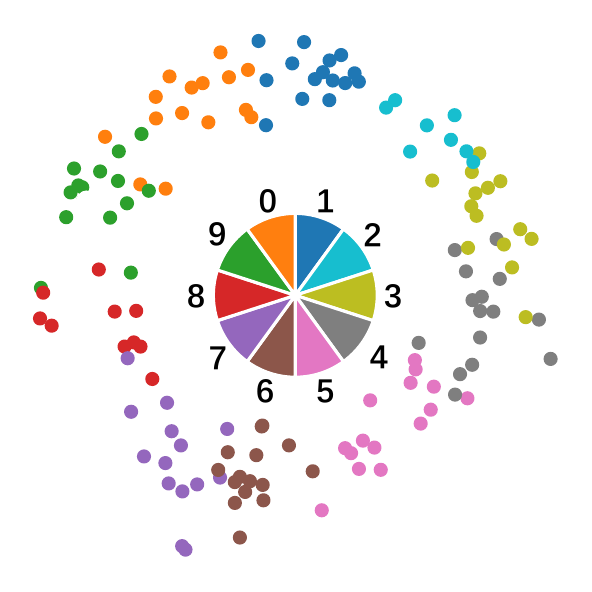}
	  }%
   \caption{Analyze the impact of the data augmentation.
   (a) shows the performance comparison on the test set with and without data augmentation. (b) and (c) are visualization results of hidden features generated by the Feedforward Block, using t-SNE. We take orientation as the feature and separate it into 10 bins. The visualization results demonstrate that by using data augmentation, the model can have a better generalization capability of orientation on the floor plane.}
   \vspace{-5mm}
\end{figure}

\subsection{Ablation Study}
\label{sec:abalation-study}
We evaluate the effectiveness of key design in \sysname, focusing on four aspects: time-frequency input (TF), block-recurrent attention module (BR), multi-loss learning (m-loss), and a detailed analysis of the impact of each loss function.

\head{Time-frequency Input}
To evaluate the effectiveness of time-frequency learning, we replaced the time-frequency dual channel with two time-domain channels.
In doing so, we observed an increase in ATE from 1.08m to 1.49m, as shown in \fig\ref{fig:ablation}.
The result confirms that representing sensory data in the time-frequency domain improves learning outcomes.

\begin{figure}[t]
\centering
   \subfloat[Building A]{%
    \label{fig:magn-a}
      \includegraphics[width=0.31\columnwidth]{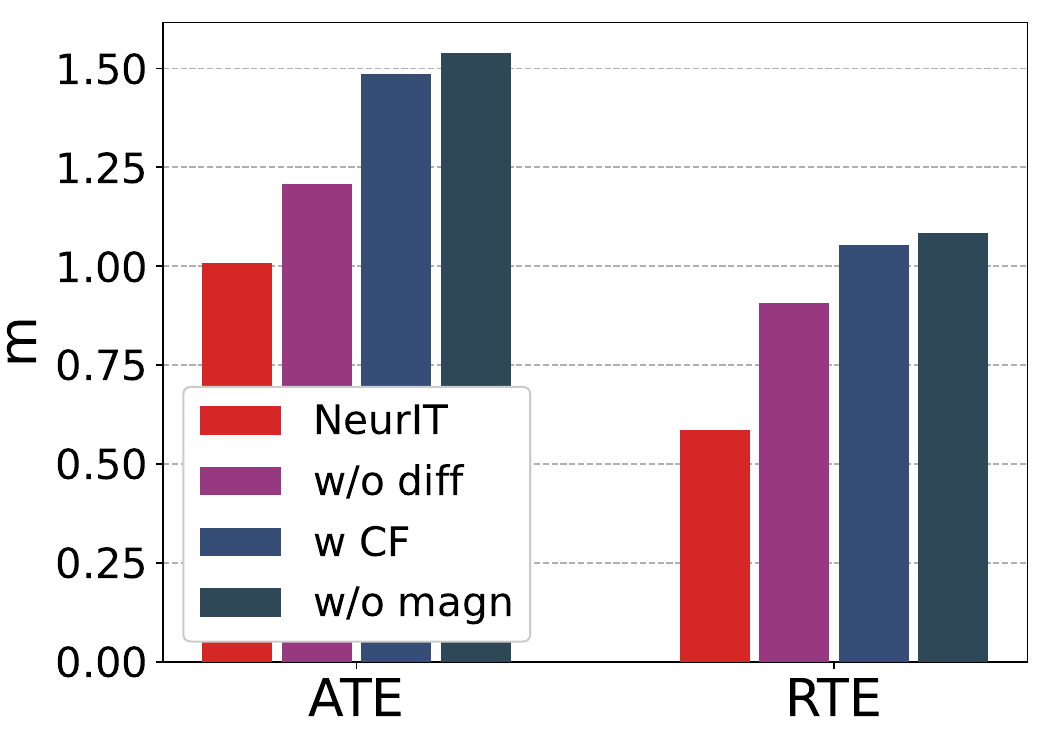}
	  }%
   \subfloat[Building B]{%
    \label{fig:magn-b}
      \includegraphics[width=0.31\columnwidth]{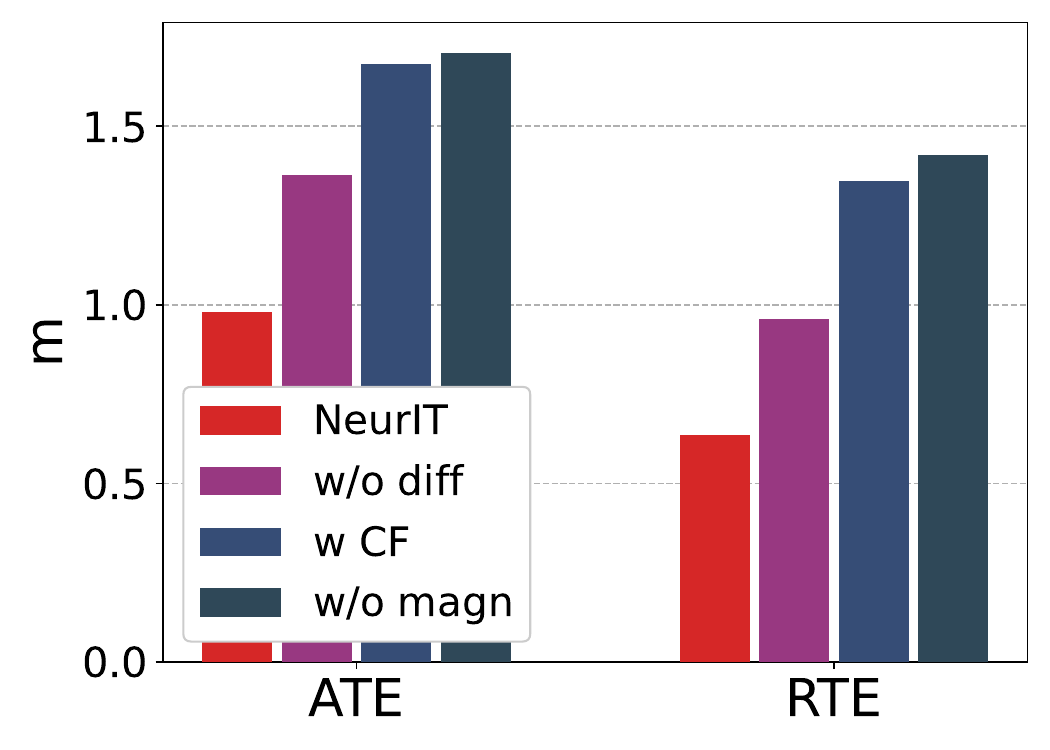}
	  }%
   \subfloat[Building C]{%
    \label{fig:magn-c}
      \includegraphics[width=0.31\columnwidth]{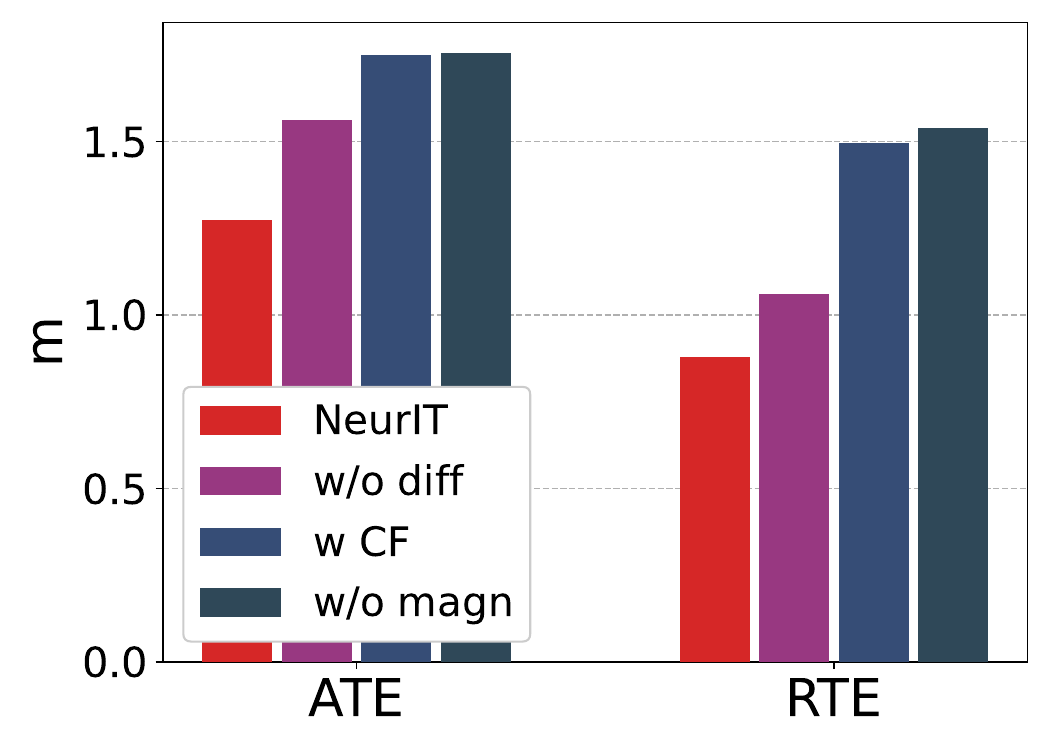}
	  }%
   \caption{Magnetometer study. \textbf{diff}: Differential operation; \textbf{CF}: Implementation of a complementary filter.}
   \label{fig:magn-study}
   \vspace{-4mm}
\end{figure}

\begin{figure}[t]
  \begin{center}
  \includegraphics[width=0.85\columnwidth]{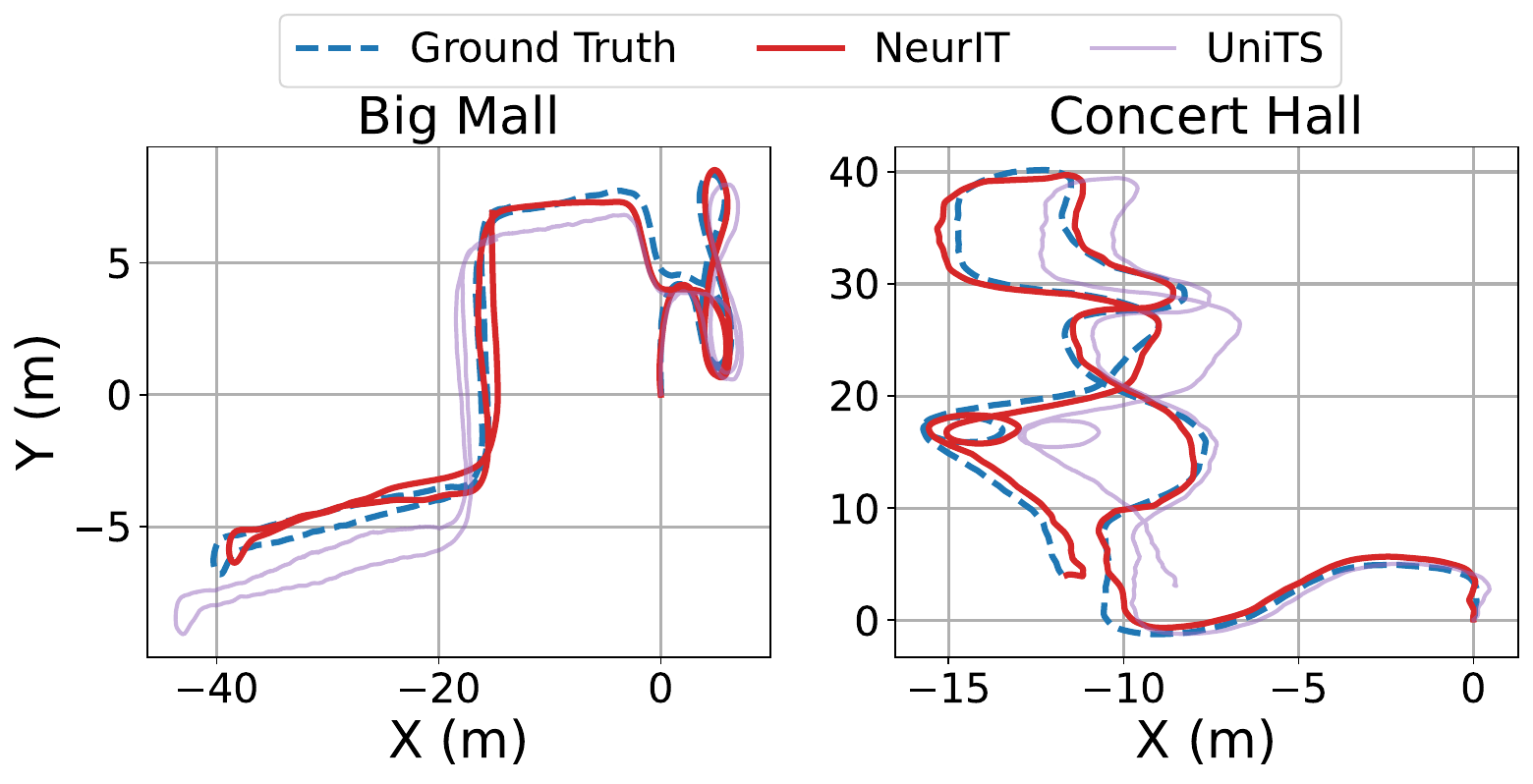}
  \caption{Examples of Predicted Trajectories in Two Large Buildings.}\label{fig:mall-exp}
  \end{center}
  \vspace{-5mm}
\end{figure}

\head{Block-recurrent Attention Module}
We replace the Block-recurrent attention module with a self-attention module to evaluate its benefits.
As shown in \fig\ref{fig:ablation}, this module contributes the most to the high performance of \sysname.
Compared to \sysname, ATE, RTE of the ablation study decreased by 52.4\%, and 94.7\%, respectively.
But even without Block-recurrent attention module, it still outperforms predictions without magnetometers.
The results validate the advantages of combining RNN with Transformer in neural inertial tracking.

\begin{table}[t]
\centering
\caption{Extended evaluation in large buildings.}
\label{tab:mall-exp}
\resizebox{0.4\textwidth}{!}{%
\begin{tabular}{c|c|cc|cc}
\hline
Metric & UniTS & \begin{tabular}[c]{@{}c@{}}NeurIT\\ w/o magn\end{tabular} & Impvt     & NeurIT & Impvt     \\ \hline
ATE    & 2.70  & 1.66                                                      & 38.41\% ↑ & 1.36   & 49.79\% ↑ \\ \hline
RTE    & 2.27  & 1.49                                                      & 34.65\% ↑ & 1.18   & 48.10\% ↑ \\ \hline
\end{tabular}%
}
\vspace{-4mm}
\end{table}

\begin{figure}[t]
\centering
   \subfloat[Comparison on ATE, RTE]{%
    \label{fig:vslam-aterte}
      \includegraphics[width=0.36\columnwidth]{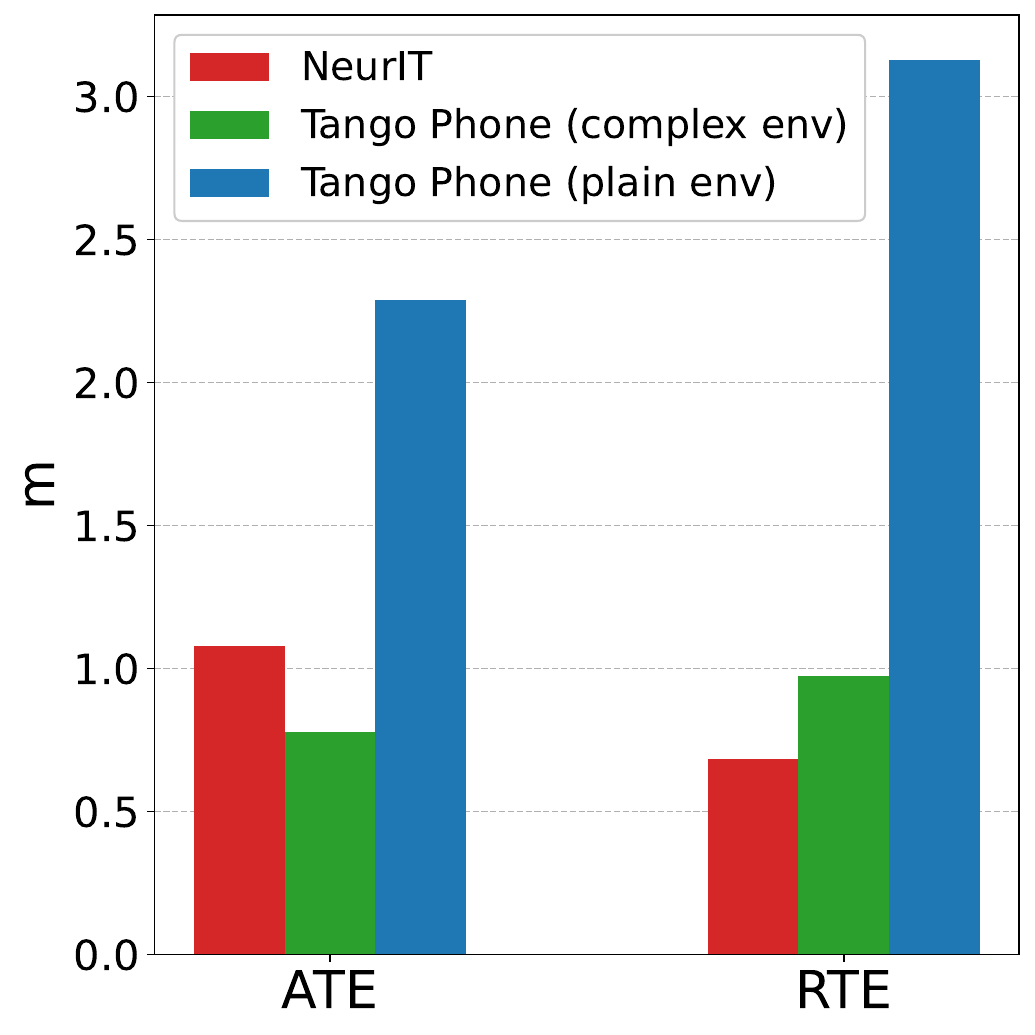}
	  }%
   \subfloat[Trajectory comparison]{%
    \label{fig:vslam-traj}
      \includegraphics[width=0.6\columnwidth]{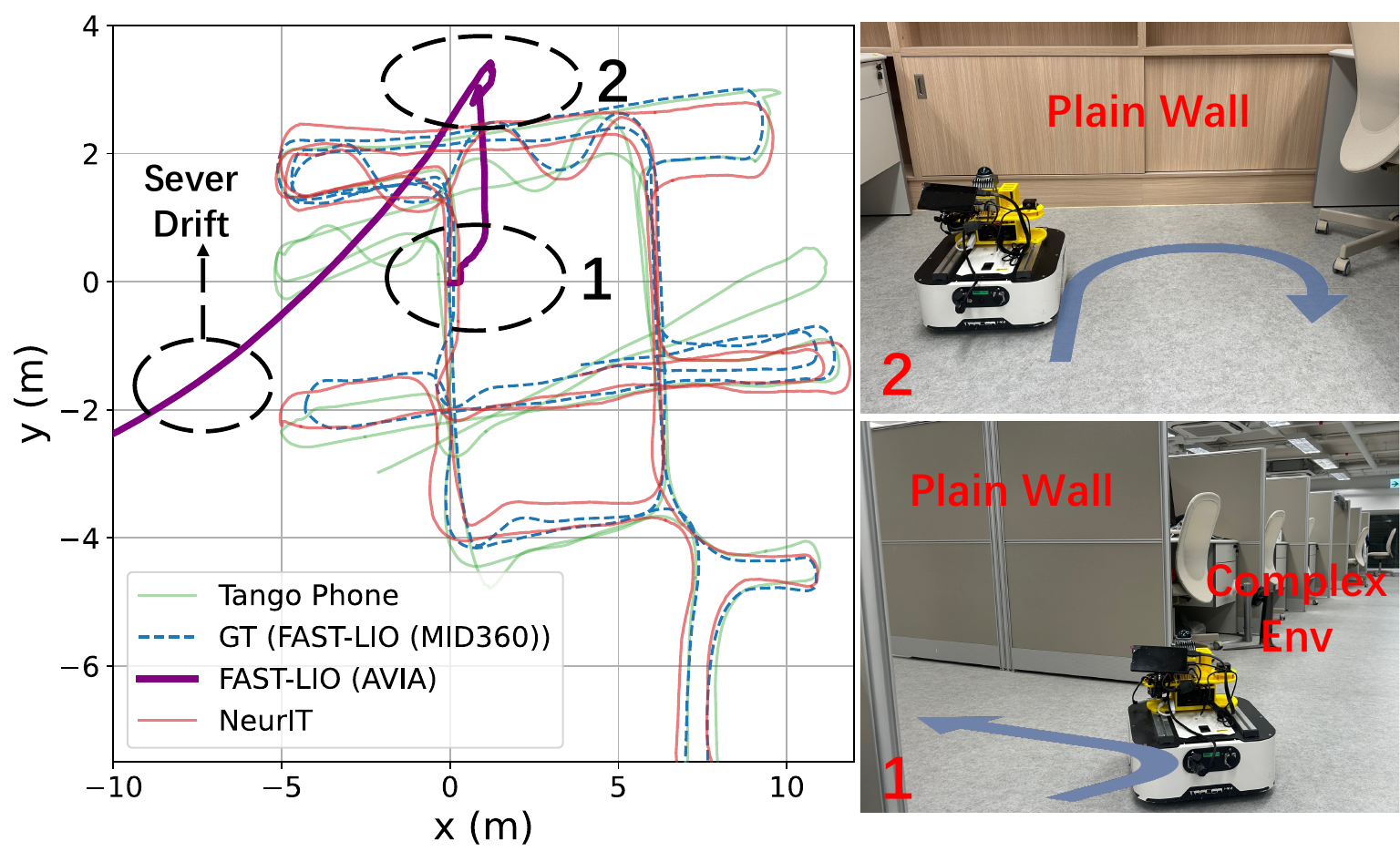}
	  }%
   \caption{Comparison with Tango Phone and narrow-FOV LiDAR (AVIA).}
   \label{fig:vslam-comp}
   \vspace{-5mm}
\end{figure}

\begin{table}[t]
\centering
\caption{Model Complexity.}
\label{tab:model-comp}
\resizebox{0.5\columnwidth}{!}{%
\begin{tabular}{l|c|c|c}
\hline
Model &
  \begin{tabular}[c]{@{}c@{}}\# Params\\ (10\textasciicircum{}6)\end{tabular} &
  \begin{tabular}[c]{@{}c@{}}Avg. GPU\\ time (ms)\end{tabular} &
  \begin{tabular}[c]{@{}c@{}}GFLOPs\\ (10\textasciicircum{}9/s)\end{tabular} \\ \hline
\sysname & 1.9   & 6.09  & 0.6   \\ \hline
IONet                   & 0.11  & 2.16  & 0.109 \\ \hline
RONIN                   & 0.22  & 4.04  & 0.107 \\ \hline
UniTS                   & 165.9 & 16.93 & 19.6  \\ \hline
\end{tabular}%
}
\end{table}

\begin{table}[t]
    \centering
    \caption{Power consumption and system complexity.}
    \label{tab:sys-evaluation}
    \resizebox{0.78\columnwidth}{!}{%
    \begin{tabular}{c|c|c|c}
    \hline
    Indicator            & Standby Mode     & GPU Mode         & CPU Mode \\ \hline
    Total Power (W)      & 3.9              & 7.1              & 4.7      \\ \hline
    GPU \& CPU Power (W) & 0.56             & 2.4              & 1.3      \\ \hline
    Avg. time (ms)       & \textbackslash{} & 34.48            & 783.4    \\ \hline
    CPU Usage (\%)       & \textbackslash{} & \textbackslash{} & 118.71   \\ \hline
    Memory Usage (Mb)    & \textbackslash{} & \textbackslash{} & 372      \\ \hline
    \end{tabular}%
    }
    \vspace{-5mm}
\end{table}

\head{Multi-loss Learning}
To evaluate the effectiveness of multi-loss learning, we first replace it with a single position loss, as used in RONIN.
This leads to a noticeable performance drop across all four metrics.
We observe some counterintuitive results: adding the orientation loss $\mathcal{L}_\mathbf{o}$ to the position loss $\mathcal{L}_\mathbf{p}$ improves heading accuracy (AYE decreases from 35° to 30°) but degrades ATE and RTE.
A similar trend is seen when combining $\mathcal{L}_\mathbf{o}$ with velocity loss $\mathcal{L}_\mathbf{v}$ while excluding $\mathcal{L}_\mathbf{p}$.
This suggests that emphasizing orientation loss may bias the model toward local rotation accuracy at the expense of global trajectory estimation.
In contrast, combining $\mathcal{L}_\mathbf{p}$ and $\mathcal{L}_\mathbf{v}$ improves overall performance.
The full multi-loss setting strikes a balance between velocity, trajectory, and orientation, leading to the most accurate trajectory predictions.

\subsection{Augmentation Study}
\label{sec:aug-study}
As discussed in \S\ref{sec:data-augmentation}, we apply data augmentation to improve generalization across all floor-plane orientations.
To validate its effectiveness, we conduct comparative experiments.
\fig\ref{fig:aug-test} shows that without augmentation, both ATE and RTE nearly double compared to the augmented setting.

To further analyze how well the model captures heading direction, we divide the 360-degree plane into 10 groups (each spanning 36°) and label the input data accordingly.
We extract the hidden embeddings from the feedforward module (before output projection) and visualize them using t-SNE~\cite{van2008visualizing} in \fig\ref{fig:aug-vis} and \fig\ref{fig:woaug-vis}.
Without augmentation, group boundaries are blurred, particularly between Groups 3 and 4, indicating poor directional discrimination.
In contrast, \fig\ref{fig:aug-vis} shows clear separation between groups, confirming that data augmentation significantly improves the model’s ability to interpret orientation-dependent IMU signals.

\subsection{Magnetometer Study}
\label{sec:magn-study}
In \sysname, we propose using \textit{differentiation of body-frame magnetometers} for orientation compensation.
To validate this component, we compare three approaches: using magnetometers without differentiation, applying a magnetometer-based complementary filter (CF) for drift compensation~\cite{shen2018closing,gong2021robust}, and using our method.
Tests on 200m+ trajectories across three buildings show our method achieves the best results (\fig\ref{fig:magn-study}).
The undifferentiated approach sees a 30\% drop, while the CF approach offers only a 4\% improvement over not using magnetometers, rendering its limitations in complex scenarios.

\subsection{Extended Experiments in Large Urban Complexes}
\label{sec:mall-exp}

To further evaluate \sysname in complex real-world settings, we conduct extended tests in two large urban environments: a shopping mall and a concert hall.
We collect 10 samples per location, each lasting approximately 5 minutes, and compare \sysname with UniTS, the strongest baseline.
As shown in \tab\ref{tab:mall-exp}, \sysname consistently outperforms UniTS on both ATE and RTE metrics, with nearly a 50\% improvement.
Our magnetometer-based compensation also yields an average accuracy gain of 15\% over models without magnetometer input.
Trajectory visualizations in \fig\ref{fig:mall-exp} highlight UniTS's drift over time, while \sysname maintains stable and accurate tracking across diverse indoor environments.

\subsection{Comparison with Tango Phone and AVIA}
\label{sec:vslam-compare}
To evaluate how \sysname compares with established visual-inertial and LiDAR-based tracking systems, we tested it against the Tango phone’s visual-inertial algorithm and the FAST-LIO algorithm using a narrow FOV LiDAR (AVIA).
In environments rich with visual features, \sysname achieves comparable performance to the Tango phone, with an ATE of 1.08m versus 0.78m.
However, in visually sparse settings, \sysname outperforms the Tango phone, which suffers from drift in featureless corners.
\fig\ref{fig:vslam-traj} illustrates the trajectory comparisons, where \sysname aligns with the ground truth.

\subsection{Training Datasize}
\label{sec:data-size}
\begin{figure}[H]
\centering
   \includegraphics[width=0.4\textwidth]{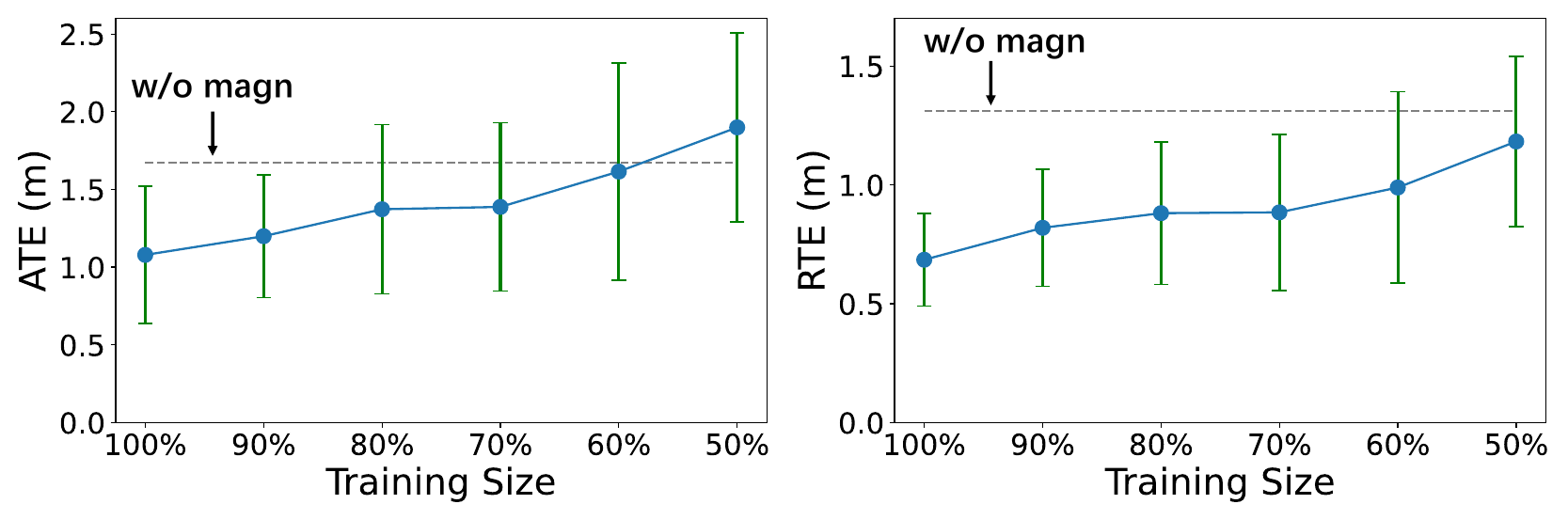}
   \vspace{-3mm}
   \caption{Analyze the impact of training size on overall performance. The gray dot line indicates the test results without taking magnetometers as input features.}
   \label{fig:train-size}
   \vspace{-5mm}
\end{figure}

We further examine how training set size affects model performance, as shown in \fig\ref{fig:train-size}.
Reducing the training set from 100\% to 50\% leads to a drop in accuracy, with ATE increasing from 1.08m to 1.89m and RTE from 0.68m to 1.18m.
Interestingly, using magnetometer data allows the model to match the performance of the non-magnetometer variant with only 60\% of the data.
Even at 50\% data, it still achieves better RTE.
These results demonstrate the robustness and strong generalization of our approach.

\subsection{System Latency}
\label{sec:system-imp}

\head{Model Complexity}
We evaluate the complexity of selected models on an NVIDIA 4090 GPU.
\tab\ref{tab:model-comp} reports each model’s parameter count, average GPU execution time for 1-second data, and GFLOPs during testing.
\sysname not only delivers the best performance but is also one of the smallest models—significantly smaller than the best baseline, UniTS.
This demonstrates that \sysname strikes a good balance between model size and computational efficiency.

\head{System Complexity} 
To examine \sysname as a real-time, power-efficient solution, we evaluate its end-to-end system complexity and power consumption on NVIDIA Jetson Orin Nano, which has been widely applied in robotic applications~\cite{alexey2021autonomous,tang2017real}, in a multi-ROS system with Intel NUC.
As shown in \tab\ref{tab:sys-evaluation}, \sysname requires 34.48ms to predict 1-second IMU data in GPU mode, including 1.62ms dedicated to coordinate system transformation and noise reduction.
The results promise \sysname as a real-time, power-efficient system to run on robotic things.

\subsection{Limitations}
\label{sec:limitation}
While \sysname consistently outperforms baseline models, our evaluation reveals two key limitations.
First, at high speeds (\textgreater 1.2,m/s), velocity prediction errors increase slightly, likely due to limited high-speed samples and increased IMU noise from wheel vibrations.
Second, long-duration trajectories show gradual drift caused by small cumulative velocity errors.
Future work could address these issues by augmenting high-speed training data, applying adaptive filtering to reduce noise, and incorporating loop-closure mechanisms to correct long-range drift in revisited indoor locations.

\section{Conclusions}
\label{sec:conclusion}
We introduce \sysname, a neural inertial tracking system that fully exploits IMU data for reliable indoor localization.
At its core is the Time-Frequency Block-recurrent Transformer (\modulename), which combines RNN and Transformer with time-frequency learning to capture both long-term dependencies and fine-grained motion patterns.
To enhance robustness, \sysname uses \textit{differentiated body-frame magnetometer} to compensate for gyroscope drift.
We deploy \sysname on a custom robotic platform and evaluate it extensively.
Results show that \sysname consistently outperforms seven state-of-the-art baselines, achieving up to 48\% higher accuracy on unseen data and maintaining an average error of one meter over 300-meter trajectories. 
These results highlight its strong generalization and real-time capability, marking a significant step toward practical, infrastructure-free indoor robotic tracking.

\section*{Acknowledgment}
This work is supported by NSFC Grant No. 62222216 and Hong Kong RGC ECS Grant No. 27204522.

\bibliographystyle{IEEEtran}
\bibliography{refs}

 




\end{document}